\documentclass{l4dc2024}

\title[Covariance-optimal MPC]{CoVO-MPC: Theoretical Analysis of Sampling-based MPC \\ and Optimal Covariance Design}

\usepackage{amsmath}
\usepackage{comment}
\usepackage{mathtools}
\usepackage{times}
\usepackage{xcolor}
\usepackage[ruled]{algorithm2e}
\usepackage{ulem}
\usepackage{hyperref}
\usepackage{xcolor}
\usepackage[capitalize]{cleveref}
\usepackage{appendix}
\usepackage{enumitem}
\usepackage{mathrsfs}
\usepackage{wrapfig}
\usepackage{xspace}
\usepackage{array}

\SetCommentSty{mycommfont}

\newboolean{isfullversion}
\setboolean{isfullversion}{True}
\newcommand{\fvtest}[2]{\ifthenelse{\boolean{isfullversion}}{#1}{#2}}
\newcommand{\hide}[1]{}

\newif\ifincludenote
\includenotefalse
\ifincludenote
    \usepackage{todonotes}
    \newcommand{\cynote}[1]{\textcolor{violet}{\textbf{Chaoyi Note:} #1}}
    \newcommand{\cytodo}[1]{\todo{\textbf{Chaoyi TODO:} #1}}

\else
    \newcommand{\cynote}[1]{}
    \newcommand{\cytodo}[1]{}
\fi
\newtheorem*{innerthm}{\innerthmname}
\newcommand{\innerthmname}{}
\newenvironment{statement}[1]
 {\renewcommand{\innerthmname}{#1}\innerthm}
 {\endinnerthm}
\newcommand{\covo}{\texttt{CoVO-MPC}\xspace}
\newcommand{\mppi}{\texttt{MPPI}\xspace}

\allowdisplaybreaks

\crefname{equation}{}{}
\crefname{figure}{Fig.}{Figs.}
\crefname{table}{Table}{Tables}
\crefname{section}{Sec.}{Secs.}
\crefname{subsection}{Sec.}{Secs.}
\crefname{subsubsection}{Sec.}{Secs.}
\crefname{appendix}{Appendix}{Appendices}
\crefname{subappendix}{Appendix}{Appendices}
\crefname{subsubappendix}{Appendix}{Appendices}
\crefname{subsubsubappendix}{Appendix}{Appendices}
\crefname{algorithm}{Alg.}{Algs.}
\crefname{theorem}{Theorem}{Theorems}
\crefname{corollary}{Corollary}{Corollaries}
\crefname{lemma}{Lemma}{Lemmas}
\crefname{definition}{Definition}{Definitions}
\crefname{remark}{Remark}{Remarks}
\newtheorem*{theorem*}{Theorem}
\newtheorem*{corollary*}{Corollary}

\author{%
 \Name{Zeji Yi}$\thanks{Equal contributions.}$ \Email{zejiy@andrew.cmu.edu }\\
 \Name{Chaoyi Pan}$^{*}$ \Email{chaoyip@andrew.cmu.edu }\\
 \Name{Guanqi He} \Email{guanqihe@andrew.cmu.edu }\\
 \Name{Guannan Qu} \Email{gqu@andrew.cmu.edu }\\
 \Name{Guanya Shi} \Email{guanyas@andrew.cmu.edu}\\
 \addr Carnegie Mellon University %
}
\begin{document}
\maketitle
\begin{abstract}%
Sampling-based Model Predictive Control (MPC) has been a practical and effective approach in many domains, notably model-based reinforcement learning, thanks to its flexibility and parallelizability. 
Despite its appealing empirical performance, the theoretical understanding, particularly in terms of convergence analysis and hyperparameter tuning, remains absent. 
In this paper, we characterize the convergence property of a widely used sampling-based MPC method, Model Predictive Path Integral Control (MPPI). We show that MPPI enjoys at least linear convergence rates when the optimization is quadratic, which covers time-varying LQR systems. We then extend to more general nonlinear systems.
Our theoretical analysis directly leads to a novel sampling-based MPC algorithm, CoVariance-Optimal MPC (\covo) that optimally schedules the sampling covariance to optimize the convergence rate. Empirically, \covo significantly outperforms standard MPPI by 43-54\% in both simulations and real-world quadrotor agile control tasks. Videos and Appendices are available at \url{https://lecar-lab.github.io/CoVO-MPC/}. 
\end{abstract}


\section{Introduction}
\label{sec:intro}
Model Predictive Control (MPC) has achieved remarkable success and become a cornerstone in various applications such as process control, robotics, transportation, and power systems~\citep{mayneRobustStochasticModel2016}. 
Sampling-based MPC, in particular, has gained significant attention in recent years due to its ability and flexibility to handle complex dynamics and cost functions and its massive parallelizability on GPUs. 
The effectiveness of sampling-based MPC has been demonstrated in various applications, including path planning
~\citep{
    helvikUsingCrossEntropyMethod2001,
    durrant-whyteCrossEntropyRandomizedMotion2012,
    nguyenTemporalPredictiveCoding2021,
    argensonModelBasedOfflinePlanning2021},
and control
~\citep{
    chuaDeepReinforcementLearning2018,
    williamsInformationTheoreticMPC2017}.
Particularly, thanks to its accessibility and flexibility to deal with learned dynamics and cost or reward functions, sampling-based MPC has been widely used as a subroutine in model-based reinforcement learning (MBRL)~\citep{
    mannorCrossEntropyMethod,
    menacheBasisFunctionAdaptation2005,
    ebertVisualForesightModelBased2018,
    zhangSOLARDeepStructured2019,
    kaiserModelBasedReinforcementLearning2020,
    bhardwajInformationTheoreticModel2020},
where the learned dynamics (often in latent space) are highly nonlinear with nonconvex cost functions. 
 
However, there is little theoretical understanding of sampling-based MPC, especially regarding its convergence and contraction properties (e.g., whether and how fast it converges to the optimal control sequence when sampled around some suboptimal control sequence).  
Moreover, despite its empirical success, there is no theoretical guideline for key hyperparameter tunings, especially the temperate $\lambda$ and the sampling covariance $\Sigma$. For instance, one of the most popular and practical sampling-based MPC algorithms, Model Predictive Path Integral Control (MPPI, \cite{williamsAggressiveDrivingModel2016,williamsInformationTheoreticMPC2017}) uses isotropic Gaussians (i.e., there is no correlation between different time steps or different control dimensions) and heuristically tunes temperature. Further, most sampling-based MPC algorithms are unaware of the underlying dynamics and optimization landscape. 
This paper makes the first step in theoretically understanding the convergence and contraction properties of sampling-based MPC. Based on such analysis, we provide a novel, practical, and effective MPC algorithm by optimally scheduling the covariance matrix $\Sigma$. Our contribution is three-fold:

\begin{itemize}[leftmargin=*,nosep]
    \item For the first time, we present the convergence analysis of MPPI. When the total cost is quadratic w.r.t. the control sequence, we prove that MPPI contracts toward the optimal control sequence and precisely characterize the contraction rate as a function of $\Sigma,\lambda$, and system parameters. We then extend beyond the quadratic setting in three cases: (1) strongly convex total cost, (2) linear systems with nonlinear residuals, and (3) general systems.   

    \item An immediate application of our theoretical results is a novel sampling-based MPC algorithm that optimizes the convergence rate, namely, CoVariance-Optimal MPC (\covo). To do so, \covo computes the optimal covariance $\Sigma$ by leveraging the underlying dynamics and cost functions, either in real-time or via offline approximations.  

    \item We thoroughly evaluate the proposed \covo algorithm in different robotic systems, from Cartpole and quadrotor simulations to real-world scenarios. In particular, compared to the standard MPPI algorithm, the performance is enhanced by $43\%$ to $54\%$ across different tasks. 
\end{itemize}

Collectively, this paper advances the theoretical understanding of sampling-based MPC and offers a practical and efficient algorithm with significant empirical advantages. This work opens avenues for further exploration and refinement of sampling-based MPC strategies, with implications for a broad spectrum of applications. 

\label{sec:related}
\vspace{-10pt}

\section{Preliminaries and Related Work}
\textbf{Notations.} In our work, \(H\) represents the horizon of the MPC. At each time step \(h\), the state is denoted as \(x_{h} \in \mathbb{R}^n\), and the control input is \(u_{h} \in \mathbb{R}^m\). The control input sequence over the horizon is \(U = u_{1:H} \in \mathbb{R}^{mH}\), which is obtained by flattening the sequence \([U]_1 = u_1, \ldots, [U]_H = u_H\). We denote random variables using the curly letter \(\mathcal{U}\), and \(U_i\) represents the \(i\)-th sample in a set of \(N\) total samples. The notation \(\xrightarrow[p]{}\), as used in our work, indicates convergence in probability.
\vspace{-10pt}

\subsection{Optimal Control and MPC}
Consider a deterministic optimal control framework:
\vspace{-9pt}
\begin{equation}
    \begin{aligned}
         & \min _{u_{1: H}} J\left(u_{1: H}\right) = \sum_{h=1}^H c_h\left(x_h, u_h\right) + c_f\left(x_{H+1}\right)  \quad  \text {s.t.} \quad x_{h+1} = f_h\left(x_h, u_h\right), \quad 1 \leq h \leq H
    \end{aligned}
    \label{eq:cost_func}
    \vspace{-6pt}
\end{equation}
Here $f_{h}: \mathbb{R}^n \times \mathbb{R}^m \rightarrow \mathbb{R}^n$ characterizes the dynamics, $c_{h}: \mathbb{R}^n \times \mathbb{R}^m \rightarrow \mathbb{R}_{\geq 0}$ captures the cost function, and $c_f: \mathbb{R}^n \rightarrow \mathbb{R}_{\geq 0}$ represents the terminal cost function.
This formulation constitutes a constrained optimization problem focused on $J(u_{1: H})$.
For a $T$-step control problem ($T \gg H$), at every step, MPC solves~\cref{eq:cost_func} in a receding-horizon manner.
In detail, upon observing the current state, MPC sets the $x_1$ in~\cref{eq:cost_func} to be the current state and solves~\cref{eq:cost_func}, which amounts to solving the optimal control only considering the costs for the next $H$ steps.
Afterward, MPC only executes the first control in the solved optimal control sequence, after which the next state can be obtained from the system, and the procedure repeats. In other words, the $H$-step problem \cref{eq:cost_func} is a subroutine for MPC. Throughout the theoretical parts (\cref{sec:problem_formulation,sec:theory}), we focus on this $H$-step subroutine.

The optimization in~\cref{eq:cost_func} is potentially highly nonconvex due to nonlinear $f_h$ and nonconvex $c_h$ and $c_f$. Non-sampling-based nonlinear MPC (NMPC) typically calls specific nonlinear programming (NLP) solvers to solve~\cref{eq:cost_func}, which depends on certain formulations like
sequential quadratic programming (SQP,~\cite{siderisEfficientSequentialLinear2005}),
differentiable dynamic programming (DDP,~\cite{tassaControllimitedDifferentialDynamic2014}),
and iterative linear quadratic regulator (iLQR,~\cite{cariusTrajectoryOptimizationImplicit2018}).
While NMPC has optimality guarantees in simple settings such as linear systems with quadratic costs, its effectiveness is limited to specific systems and solvers~\citep{songReachingLimitAutonomous2023,forbesModelPredictiveControl2015}. In the next section, we introduce sampling-based MPC, a recently popular paradigm that can mitigate these limitations.

\vspace{-5pt}
\subsection{Sampling-based MPC and Applications in Model-based Reinforcement Learning} \label{subsec:samplingmpc}
Recently, sampling-based MPC gained popularity as an alternative to NMPC by employing the zeroth-order optimization strategy~\citep{drewsVisionBasedHighSpeed2018, mannorCrossEntropyMethod, helvikUsingCrossEntropyMethod2001}. 
Instead of deploying specific NLP solvers to optimize~\cref{eq:cost_func}, sampling-based MPC samples a set of control sequences from a particular distribution, evaluates the costs of all sequences by rolling them out, optimizes and updates the sampling distribution as a function of samples and their evaluated costs.
Unlike NMPC, sampling-based MPC has no particular assumptions on dynamics or cost functions, resulting in a versatile approach in different systems. 
Moreover, the sampling nature enables massive parallelization on modern GPUs.
For example, MPPI is deployed onboard for real-time robotic control~\citep{williamsAggressiveDrivingModel2016,pravitraL1AdaptiveMPPIArchitecture2020,sacksDeepModelPredictive2023}, sampling thousands of trajectories at more than 50Hz with a single GPU. 

Sampling-based MPC can be categorized by what sampling-based optimizer it deploys. Covariance Matrix Adaptation Evolution Strategy (CMA-ES,~\cite{hansenReducingTimeComplexity2003}) adapts sampling mean and covariance using all samples to align with the desired distribution. Cross-Entropy Method (CEM,~\cite{botevChapterCrossEntropyMethod2013}) deploys a similar procedure but only considers the low-cost samples.
\cite{wagenerOnlineLearningApproach2019} unifies different sampling-based MPC approaches by connecting them to online learning with different utility functions and Bregman divergences.
Among these variants, MPPI~\citep{williamsInformationTheoreticMPC2017,williamsAggressiveDrivingModel2016} is one of the most popular and practical, and it reformulates the control problem into optimal distribution matching from an information-theoretic perspective.
MPPI's performance is sensitive to hyperparameters and system dynamics, leading to several variants that improve its efficiency and robustness, e.g., using Tsallis Divergence~\citep{wangVariationalInferenceMPC2021}, imposing extra constraints when sampling~\citep{gandhiRobustModelPredictive2021, balciConstrainedCovarianceSteering2022}, or using learned optimizers~\citep{sacksDeepModelPredictive2023}.
Compared to these works, \covo optimally schedules the sampling covariance followed by our rigorous theoretical analysis.

MPPI has been particularly popular in the Model-based Reinforcement Learning (MBRL) context, an RL paradigm that first learns a dynamics model and optimizes a policy using the learned model. 
Specifically, largely because of its parallelizability and its flexibility to handle nonlinear and potentially latent learned models, MPPI has been a popular control and planning subroutine in MBRL~\citep{menacheBasisFunctionAdaptation2005, LearningTetrisUsing, jannerWhenTrustYour2021, lowreyPlanOnlineLearn2019, hafnerLearningLatentDynamics2019}.
Moreover, MPPI and CEM have been widely deployed for uncertainty-aware control and planning with probabilistic models (e.g., PETS~\citep{chuaDeepReinforcementLearning2018}, PlaNet~\citep{hafnerLearningLatentDynamics2019}), and integrated with learned value and policy to improve the MBRL performance (e.g., TD-MPC~\citep{hansenTemporalDifferenceLearning2022}, MBPO~\citep{argensonModelBasedOfflinePlanning2021}).

Despite its broad applications, there is little theoretical understanding of sampling-based MPC algorithms. In this paper, we aim to provide theoretical characterizations of MPPI's optimality and convergence, based on which we design an improved algorithm \covo. In addition, \covo can potentially be an efficient backbone for a broad class of MBRL algorithms.

\vspace{-8pt}
\section{Problem Formulation}
\label{sec:problem_formulation}

While MPPI is an iterative approach with a receding horizon, to streamline the analysis, we consider a single step in MPPI, that is solving a trajectory optimization problem~\cref{eq:cost_func} using sampling-based method\footnote{We will discuss the full implementation of MPPI with a receding horizon in~\cref{sec:algo}.}.
Specifically, MPPI rewrites the cost function in~\cref{eq:cost_func}, as only a function of the control input (i.e., substitute the state $x_h$ with control inputs $U$). To minimize $J(U)$, MPPI samples the control sequence $U$ and outputs a weighted sum of these samples. Specifically, the control sequence is sampled from a Gaussian distribution: $ \mathcal{U} \sim \mathcal{N}(U_{\mathrm{in}},\Sigma)$, where $U_{\mathrm{in}} \in \mathbb{R}^{mH}$ is the mean and $\Sigma \in \mathbb{R}^{mH\times mH}$ is the covariance matrix. Given the samples, we calculate the output control sequence $U_{\mathrm{out}}\in \mathbb{R}^{mH}$ using a softmax-style weighted sum based on each sample's cost $J(U_i)$:
\vspace{-2pt}
\begin{equation}
    \small
    U_{\mathrm{out}} = \frac{\sum_{i=1}^{N} U_i \exp\left(-\frac{J(U_i)}{\lambda}\right)}{\sum_{i=1}^N \exp\left(-\frac{J(U_i)}{\lambda}\right)},
    \label{eq:MPPI}
\end{equation}
where $\lambda$ is the temperature parameter commonly used in statistical physics~\citep{busetti2003simulated} and machine learning.
As $\lambda\rightarrow 0$, the focus intensifies on the top control sequence, while a larger $\lambda$ distributes the influence over a set of relatively favorable trajectories. In the next section, we will characterize the single-step convergence properties of MPPI in different systems.
\vspace{-10pt}

\section{Main Theoretical Results}
\label{sec:theory}
\vspace{-5pt}
Our primary goal is to investigate the optimality and convergence rate of MPPI.
We are driven by the following questions:
Under what conditions does the algorithm exhibit outputs close to the optimal solution?
And what is the corresponding convergence rate?
Moreover, if the convergence is influenced by factors such as \(U_{\mathrm{in}}, \lambda, \Sigma\), how can we strategically design the sample covariance to expedite the convergence process?

To answer the above questions, we first analyze the convergence of the single-step MPPI~\cref{eq:MPPI} in a quadratic cost's environment.
In \cref{sec:quad_conv}, we show that the expected result of equation~\cref{eq:MPPI} contracts to the optimal solution in \cref{theo:1_contraction}.
Then, in \cref{sec:opt_design}, based on the contraction of the expectation output, we further give an optimal design principle for the covariance to better utilize the known dynamics. Moreover, in \cref{subsec:non_linear}, we address general nonlinear environments and prove that MPPI still keeps the contraction property, though with bounded associated errors.

\vspace{-8pt}
\subsection{Convergence Analysis for Quadratic \texorpdfstring{$J(U)$}{J(U)}}
\label{sec:quad_conv}
We start by considering the total cost $J(U)$ of the following form,
\begin{equation}
    \label{eq:quad_cost}
    J(U) = U^\top D U + U^\top d,
\end{equation}

which is quadratic in $U$. One simple but general example that satisfies the above is when the dynamics is LTI or LTV, and the cost is quadratic, which we will elaborate on in \cref{ex:LQR}. We will also discuss the generalization to non-quadratic $J(U)$ and nonlinear dynamics in \cref{subsec:non_linear}.

In this setting, our first result proves the convergence of MPPI, i.e., \cref{eq:MPPI}. Specifically, we show that the expected output control sequence contracts, and critically, we provide a precise contraction rate in terms of the system parameter $D$ and the algorithm parameters $\Sigma,\lambda$.

\begin{theorem}
    Given $U_{\mathrm{in}}$ and the sampling distribution $\mathcal{N}(U_{\mathrm{in}},\Sigma)$, under the assumption that the total cost $J(U)$ is in quadratic form in \cref{eq:quad_cost} and $D$ is positive semi-definite, the weighted sum of the samples $U_{\mathrm{out}}$, as in \cref{eq:MPPI}, converges in probability to a contraction towards the optimal solution $U^*$ when the number of sample $N \rightarrow \infty$ in the following way:
    \begin{equation}
        \frac{||U_{\mathrm{out}} - U^*||}{||U_{\mathrm{in}} - U^*||}  \xrightarrow[p]{} \left\| (\frac{2}{\lambda}\Sigma D+I)^{-1} \right\| < 1
        \label{eq:expectation_contraction_u}
    \end{equation}
    Similarly, the expected cost contracts to the optima $J^*$ in the following way:
    \begin{equation}
        J(U_{\mathrm{out}}) - J^*\xrightarrow[p]{} J_{\mathrm{out}} - J^*\leq (J(U_{\mathrm{in}})-J^*)\left\|(I+\frac{2}{\lambda}D^{\frac{1}{2}}\Sigma D^{\frac{1}{2}})^{-2} \right\|,
        \label{eq:expectation_contraction}
    \end{equation}
    where $J_{\mathrm{out}}$ is a constant determined by function J and $U_{\mathrm{in}},\Sigma,D$.
    \label{theo:1_contraction}
\end{theorem}

\Cref{theo:1_contraction} means, when the number of sample $N$ goes to infinity, MPPI enjoys a linear contraction\footnote{MPPI can be considered a regularized Newton's method, utilizing the Hessian Matrix $D$, which typically results in linear convergence (contraction).
    Moreover, it exhibits superlinear convergence in the proximity of the optimum.
    Further discussion on this topic will be provided in \fvtest{\cref{apd:Quad_J}}{\href{https://tinyurl.com/covo-mpc-cmu}{Appendix} A\hide{~\citep{zejiCoVOMPCTheoreticalAnalysis2023}}}.} towards the optimal control sequence $U^*$ and the optimal cost $J^*$.
Further, from the contraction rate in both \cref{eq:expectation_contraction_u} and \cref{eq:expectation_contraction}, we can observe that smaller $\lambda$ leads to faster contraction, which we will discuss more on in \cref{subsec:non_linear}.
However, the faster contraction comes at the cost of higher sample complexity due to a higher variance on $U_{\mathrm{out}}$, as shown by \citet{busetti2003simulated}.

In addition to $\lambda$, the contraction rate is also ruled by the matrix $D^{\frac{1}{2}}\Sigma D^{\frac{1}{2}}$, indicating that choosing $\Sigma$ is vital for the contraction rate, and further, the optimal choice of $\Sigma$ should depend on the Hessian matrix $D$.
Note that the standard MPPI chooses $\Sigma = \sigma I$, i.e., sampling isotropically over a landscape defined by $D$, which could yield slow convergence.
Based on this observation, we will develop an optimal design scheme of $\Sigma$ in \cref{sec:opt_design}.

\begin{example}[Time-variant LQR]
    \label{ex:LQR}
    Here is an example of the linear quadratic regulator (LQR) setting with time-variant dynamics and costs.
    The dynamics is $x_{h+1} = A_h x_h + B_h [U]_h + w_h$ and the cost is given by
    $
        J(U)=\sum_{h=1}^{H}\left(x_h^\top Q_h x_h+[U]_h^\top R_h [U]_h\right)
    $, where $U=\left([U]_1, \ldots, [U]_{H}\right)$ and $Q_h\succeq 0, R_h \succ 0$. Plugging in the states $x_h$ as a function of $U$, the cost can be reformulated as $J(U) = U^\top D U + U^\top d + J_0$ where $d$ and $J_0$ are constants derived from system parameters. And the Hessian is \( D = M^\top \mathcal{Q} M + \mathcal{R} \), with
    $$
        \small
        \label{eq:hessian}
        M =
        \begin{bmatrix}
            0                       & \cdots                &        &         \\
            B_1                     & 0                     & \cdots &         \\
            \Tilde{A}_{2,2} B_{1}   & B_2                   & 0      & \cdots  \\
            \vdots                  & \vdots                &        &         \\
            \Tilde{A}_{H-1,2} B_{1} & \Tilde{A}_{H-1,3} B_2 & \cdots & B_{H-1}
        \end{bmatrix},
        \mathcal{Q} =
        \begin{bmatrix}
            Q_1    & 0      & \cdots &     \\
            0      & Q_2    & \cdots &     \\
            \vdots & \vdots &        &     \\
            0      & 0      & \cdots & Q_H
        \end{bmatrix},
        \mathcal{R} =
        \begin{bmatrix}
            R_1    & 0      & \cdots &     \\
            0      & R_2    & \cdots &     \\
            \vdots & \vdots &        &     \\
            0      & 0      & \cdots & R_H
        \end{bmatrix},
    $$
    and $\Tilde{A}_{i,j} = A_i A_{i-1} \cdots A_j$ for $i \geq j$.
\end{example}
\vspace{-5pt}

\cref{theo:1_contraction} primarily addresses scenarios with quadratic $J(U)$, suitable for time-varying LQR as seen in \Cref{ex:LQR}.
Even though time-varying LQR can be solved using well-known analytical methods, our theoretical exploration still holds significant value, and its underlying principles extend beyond linear systems.
Firstly, implementation-wise, nonlinear systems can be linearized around a nominal trajectory, resulting in a Linear Time-Variant (LTV) system to which \cref{theo:1_contraction} directly applies, like in iterative Linear Quadratic Regulator (iLQR, \cite{ITERATIVELINEARQUADRATIC2004}) and Differential Dynamic Programming (DDP, \cite{mayne1966second}).
Secondly, from a theory perspective, we further discuss the linearization error and more generally, non-quadratic $J(U)$ in \cref{subsec:non_linear}. Lastly, the analysis within this quadratic framework elucidates the optimal covariance design principle in \cref{sec:opt_design}, which we use to design the \covo algorithm in \cref{sec:algo} that can be implemented beyond linear systems. Empirically (\cref{sec:exp}), \covo demonstrates superior performance across various nonlinear problems, affirming the value of \cref{theo:1_contraction} beyond quadratic formulations.

\vspace{-5pt}
\subsection{Optimal Covariance Design}
\label{sec:opt_design}

\cref{theo:1_contraction} proves that MPPI guarantees a contraction to the optima with contraction rate depending on the choice of $\Sigma$ (see~\cref{eq:expectation_contraction}).
We now investigate the optimal $\Sigma$ to achieve a faster contraction rate. Based on \cref{eq:expectation_contraction}, it is evident that scaling $\Sigma$ with a scalar larger than $1$ brings better contraction in expectation.
However, doing so is equivalent to decreasing $\lambda$, which will lead to higher sample complexity according to \cref{sec:quad_conv}.
In light of this, we formulate the optimal covariance $\Sigma$ design problem as a constrained optimization problem: How can we design $\Sigma$ to achieve an optimal contraction rate, subject to the constraint that the determinant of the covariance is upper-bounded, that is $\det \Sigma  \leq \alpha$. The following theorem optimally solves this constrained optimization problem.
\vspace{-5pt}
\begin{theorem}
    \label{theo:optimal_sigma} Under the constraint that positive semi-definite matrix $\Sigma$ satisfies $\det \Sigma  \leq \alpha$, the contraction rate $\left\|(I+\frac{2}{\lambda}D^{\frac{1}{2}}\Sigma D^{\frac{1}{2}})^{-2} \right\|$ in \cref{eq:expectation_contraction} is minimized when $\Sigma$ has the following form: The Singular Value Decomposition (SVD) of $\Sigma = V^\top \Lambda V$ shares the same eigenvector matrix $V$ with the SVD of the Hessian matrix $D = V^\top \mathcal{O} V$, and
    \begin{equation}
        \frac{[\mathcal{O}]_{i}^2 [\Lambda_{\mathrm{in}}]_i^2 [\Lambda]_{i}}{(1+\frac{2}{\lambda}[\Lambda]_{i} [\mathcal{O}]_{i})^3}  = \mathrm{Constant}, \forall i=1,\cdots,mH,
        \label{eq:opt_cov_exact}
    \end{equation}
    where $\Lambda,\mathcal{O}\in \mathbb{R}^{mH \times mH}$, $\Lambda = diag([\Lambda]_1,[\Lambda]_2,\ldots,[\Lambda]_{mH}])$,$ \mathcal{O} = diag([\mathcal{O}]_1,[\mathcal{O}]_2,\ldots,[\mathcal{O}]_{mH})$. Further, $\Lambda_{\mathrm{in}}\in \mathbb{R}^{mH}$is the coordinates of $U_{\mathrm{in}} - U^*$ under the basis formed by the eigenvectors $V$. In other words, $ \Lambda_{\mathrm{in}} = V^\top (U_{\mathrm{in}} - U^*)= [[\Lambda_{\mathrm{in}}]_1,[\Lambda_{\mathrm{in}}]_2,\ldots,[\Lambda_{\mathrm{in}}]_{mH}]^\top$. Lastly, the $\mathrm{Constant}$ in the right-hand-side of \cref{eq:opt_cov_exact} is selected subject to the constraint $\det \Sigma=\alpha$.
\end{theorem}

\vspace{-5pt}
While \cref{theo:optimal_sigma} offers insights into the optimal design of $\Sigma$, solving \cref{eq:opt_cov_exact} relies on prior knowledge of $U^*$. In \cref{coro:sigma} below, we present an approximation to the solution of \cref{eq:opt_cov_exact} in \cref{theo:optimal_sigma}. This approximation is not only efficient to solve but also eliminates the need for knowing $U^*$.
These advantages come under the assumptions of an isotropic gap between $U^*$ and $U_{\mathrm{in}}$ and a small $\lambda$ regime. 
The isotropy gap assumption is necessary as we do not have additional information about $U^*$\footnote{We show that the isotropy assumption is minimizing the max cost from a family of gap $U_{\mathrm{in}}-U^*$ in \fvtest{Remark \ref{rem:apd_opt_cov} of \cref{apd:opt_cov}}{the \href{https://tinyurl.com/covo-mpc-cmu}{Appendix} A\hide{~\citep{zejiCoVOMPCTheoreticalAnalysis2023}}}.}.
A small $\lambda$ regime is also practical since \cref{theo:1_contraction} implies a smaller $\lambda$ yields faster contraction given a sufficiently large sample number $N$.
Given modern GPU's capabilities in large-scale parallel sampling, we can generate sufficient samples for small values of $\lambda$ (in practice, MPPI often uses $\lambda < 0.01$), which justifies the adoption of the small $\lambda$ regime in~\cref{coro:sigma}. Notably, this approximation serves as the foundational design choice for \covo introduced in~\cref{sec:algo}, and empirical experiments consistently validate its feasibility.

\vspace{-5pt}
\begin{corollary}\label[corollary]{coro:sigma}
    Under the assumption $U_{\mathrm{in}} - U^* \sim \mathcal{N}(0,I)$, when $\lambda \rightarrow 0$, asymptotically, the optimal solution of \cref{eq:opt_cov_exact} is:
    \begin{equation}
        \label{eq:optimal_sigma}
        \lim_{\lambda\rightarrow0,N \rightarrow \infty} \Sigma = (\alpha \det D^{1/2})^{\frac{1}{mH}} D^{-1/2}
    \end{equation}
\end{corollary}
\vspace{-4pt}

Note that in both \cref{theo:optimal_sigma} and \cref{coro:sigma}, the choice of $\Sigma$ depends on system parameters. The underlying concept of this design principle is to move away from isotropic sampling of the control sequence. Instead, the sample distribution is tailored to align with the structure of the system dynamics and cost functions. In essence, if we view the covariance matrix as an ellipsoid in the control action space, \cref{theo:optimal_sigma} and \cref{coro:sigma} implies that the optimal covariance matrix should be compressed in the direction where the total cost $J(U)$ is smoother and stretched along the direction aligned with the $J$'s gradient. Guided by this optimal covariance matrix design principle, we introduce a practical algorithm in \cref{sec:algo} that incorporates this optimality into the MPPI framework. This framework computes the Hessian matrix and designs the covariance matrix, thus leveraging the system structure for improved performance.

\vspace{-5pt}
\subsection{Generalization Beyond the Quadratic \texorpdfstring{$J(U)$}{J(U)}}
\label{subsec:non_linear}

\cref{sec:quad_conv} gives the convergence properties of MPPI when $J(U)$ is quadratic, but in practice, MPPI works well beyond quadratic $J(U)$.
Therefore, in this section, we extend the convergence analysis in \cref{sec:quad_conv} to general non-quadratic total costs. 
Specifically, we consider three different settings, from strongly convex (not necessarily quadratic) $J(U)$, linear systems with nonlinear residuals, to general systems.

In many cases, the total cost function \(J\) demonstrates strong convexity without necessarily being quadratic. A typical example is the time-varying LQR (\cref{ex:LQR}) with the quadratic term \( x^\top Q x \) in the cost being replaced with a strongly convex function. In light of this, we now study the convergence of MPPI under a \( \beta \)-strongly convex total cost function \( J(U) \).

\vspace{-5pt}
\begin{theorem}\label{theo:small_lambda}\textnormal{(Strongly convex $J$. Informal version of \fvtest{\cref{theo:small_lambda} in \cref{apd:proof_str_conv}}{Theorem 4 in \href{https://tinyurl.com/covo-mpc-cmu}{Appendix} A.3})}
    Given a $\beta$-strongly convex function $J(U)$ which has Lipschitz continuous derivatives, that $\frac{\partial J}{\partial U} $ is $ L_d$-Lipschitz,  $U_{\mathrm{out}}$ converges to a neighborhood of the optimal point $U^*$ in probability as the number of samples $N$ tends to infinity, i.e., $U_{\mathrm{out}} \xrightarrow[p]{} U_{c}$. Here $U_c$ satisfies $\left\|U_{c} - U^*\right\|   \leq  \frac{\lambda}{\beta}\frac{2\left\|  \Sigma^{-1} \right\|}{(1+\frac{\lambda}{\beta \left\|\Sigma\right\|})}\left\| U_{\mathrm{in}}-U^{*}\right\| + \left\| U_{\mathrm{error}} \right\|$, where $\left\|U_{\mathrm{error}}\right\| \sim O(\frac{\sqrt{\lambda}}{\beta})$.
    
\end{theorem}
\vspace{-5pt}

\cref{theo:small_lambda} demonstrates MPPI maintains a contraction rate of $O(\frac{\lambda}{\beta})$ with a small $\lambda$, approaching $0$ when $\lambda \ll \beta$. The remaining residual error $U_{\mathrm{error}}$, also of $O(\frac{\lambda}{\beta})$, suggests MPPI's convergence to a neighborhood around the optimal point with size $O(\frac{\lambda}{\beta})$. The contraction rate in strongly convex $J(U)$ differs from the quadratic case (\cref{sec:quad_conv}) due to the difference between the quadratic density function in Gaussian. The non-quadratic $J(U)$ has a slightly looser bound in terms of constant. 

Beyond strongly convex $J(U)$, we next consider nonlinear dynamics. 
Specifically, we consider the same cost function as in \cref{ex:LQR} with the following dynamics: $x_{h+1} = A x_h+B [U]_h + w + g([U]_h)$, where \(w\) is a constant, and \(g\) represents nonlinear residual dynamics. 
We then recast the total cost as $J(U) = U^\top D U + d^\top U +J_{\mathrm{res}}(U)$ by retaining all higher-order terms in $J_{\mathrm{res}}(U)$. 
Subsequently, we establish an exponential family using \( D \), \( d \), and \( J_{\mathrm{res}}(U) \) as sufficient statistics. 
Given the characteristics of the residual dynamics, we can bound the variations in these three statistics relative to the quadratic total cost. 
Utilizing the properties of the exponential family, we then show that, with bounded residual dynamics $g$,  contraction is still guaranteed to some extent.

\vspace{-5pt}
\begin{theorem}
\textnormal{(Linear systems with nonlinear residuals. \fvtest{\footnotesize{Informal version of \cref{inf:lip} in \cref{apld:proof-res}}\normalsize}{Informal version of Theorem 5 in \href{https://tinyurl.com/covo-mpc-cmu}{Appendix} A.3})}.
    Given the residual dynamics $g$, $U_{\mathrm{out}}$ contracts in the same way as in \cref{eq:expectation_contraction_u} with contraction error $\left\|U_{\mathrm{error}}\right\|\sim \frac{L_1}{\lambda}\left\| Q \right\|(O(\left\| A^{3H}\right\|)C_3 + O(\left\|A^{2H}\right\|)C_2)$ with $L_1$ as a Lipschitz constant from the exponential family, and $C_2,C_3$ as constants that coming from the bounded residual dynamics $g$.
    \label{inf:lip}
\end{theorem}
\vspace{-5pt}

Additionally, it is also worthnoting that the Lipschitz constant \( L_1 \) is task-specific since different tasks have different sufficient statistics \( J_{\mathrm{res}} \) within the exponential family. Given \cref{inf:lip} for residual dynamics, a direct and straightforward corollary arises for general systems, with direct access to the cost function, specifically \( J_{\mathrm{res}} \).

\vspace{-5pt}
\begin{corollary}\textnormal{(General systems. Informal version of \fvtest{\cref{cor:general_nonlinear} in \cref{apd:proof_general}}{Theorem 6 in \href{https://tinyurl.com/covo-mpc-cmu}{Appendix} A.3}.)}
    $U_{\mathrm{out}}$ contracts the same as \cref{eq:expectation_contraction_u} with a contraction error, $\left\|U_{\mathrm{error}}\right\|\sim O(\frac{L_1}{\lambda}(\left\|D -D'\right\|))$, where $D'$ is the Hessian under the residual dynamic $g$, and $D$ is the Hessian without the residual dynamic.
    \label{cor:general_nonlinear}
\end{corollary}
\vspace{-10pt}

\section{The CoVariance-Optimal MPC (CoVO-MPC) Algorithm}
\label{sec:algo}
\setlength{\intextsep}{0pt}
\setlength{\columnsep}{18pt}
\begin{wrapfigure}{r}{0.5\textwidth}
    \begin{algorithm2e}[H]
        \DontPrintSemicolon
        \LinesNumbered
        \SetAlgoLined
        \caption{CoVO-MPC}
        \label{alg:covo}
        \KwIn{$H$;
            $N;$
            $T;$
            $c_{1:T};$
            $\mathbf{C}(\cdot);$
            $\texttt{shift}(\cdot);$}
        \BlankLine
{\small
        \For{$t=1:T$}{
        $\Sigma_{t} \gets \mathbf{C}(D_{t} = \nabla^2 J_t(U_{\mathrm{in}|t}))$ \label{line:covariance}
            
        Sample $U_{i|t} \sim \mathcal{N}(U_{\mathrm{in}|t}, \Sigma_{t})$ for $1 \leq i \leq N$ \label{line:sample}
        
        Compute weight $\kappa_{i|t} \gets \frac{\exp(-\frac{J_t(U_{i|t})}{\lambda})}{\sum_{i=1}^N \exp(-\frac{J_t(U_{i|t})}{\lambda}))}$ \label{line:weight_cost}
        
        $U_{\mathrm{out}|t} \gets \sum_{i=1}^N \kappa_{i|t} U_{i|t}$ \label{line:sum}
        
        Execute $u_t \gets [U_{\mathrm{out}|t}]_1$, receive $x_{t+1}$ \label{line:control}

        $U_{\mathrm{in}|t+1} \gets \texttt{shift}(U_{\mathrm{out}|t})$ \label{line:shift}
        }
        }
    \end{algorithm2e}
\end{wrapfigure}

\cref{coro:sigma} in \cref{sec:opt_design} provides the optimal covariance matrix $\Sigma$ as a function of the Hessian matrix $D$. We define the mapping in \cref{eq:optimal_sigma} as $\Sigma = \mathbf{C}(D)$. Deploying $\mathbf{C}(\cdot)$ at every time step in the sampling-based MPC framework, we propose \covo (\cref{alg:covo}), a general sampling-based MPC algorithm. 

As shown in~\cref{line:covariance} of~\cref{alg:covo}, at each time step, \covo will first generate an optimal sampling covariance matrix $\Sigma_t$ based on the cost's Hessian matrix around the sampling mean \(\nabla^2 J_t(U_{\mathrm{in}|t})\).
Next, we will follow the MPPI framework to calculate control sequences from the sampling distribution (\cref{line:sample,line:weight_cost,line:sum}) and execute the first control command in the sequence (\cref{line:control}).
After that, we will shift the sampling mean forward to the next time step using the $\texttt{shift}$ operator (\cref{line:shift}) and repeat the process.

The optimal covariance design (\cref{line:covariance}) is critical for \covo, which requires computing the Hessian matrix $D_t=\nabla^2 J_t(U_{\mathrm{in}|t}))$ in real-time. 
The total cost $J_t(U_{\mathrm{in}|t}) = \sum_{h=1}^{H} c_{t+h}(x_{t+h|t}, [U_{\mathrm{in}|t}]_h)$ is gathered by rolling out the sampling mean sequence $U_{\mathrm{in}|t}$ with the dynamics initialized at $x_t$, where $x_{t+h|t}$ is the $h^\mathrm{th}$ rollout state and $c_t$ is the running cost at time step $t$.
To ensure \(D_t\) to be positive definite, we add a small positive value to the diagonal elements of $D_t$ such that $D_t \succeq \epsilon I \succ 0$, which is equivalent to adding a small quadratic control penalty term to the cost function.

\paragraph{Offline covariance matrix approximation.} 
Computing $D_t$ and $\Sigma_t$ in real time could be expensive. Therefore, we propose an offline approximation variant of \covo. 
In this variant, we cache the covariance matrices across all $t$ offline by rolling out the dynamics using a nominal controller (e.g., PID). 
More specifically, offline, we calculate the whole sequence of covariance matrix $\Sigma^\text{off}_{1:T}$ in simulation using the nominal controller. Then, online, $\Sigma^\text{off}_{t}$ serves as an approximation of the optimal covariance matrix. The details can be found in~\fvtest{\cref{sec:offline}}{\href{https://tinyurl.com/covo-mpc-cmu}{Appendix} B.1\hide{ \citep{zejiCoVOMPCTheoreticalAnalysis2023}}}. Empirically, we observe that this offline approximation performs marginally worse than \covo with less computation. 

\vspace{-10pt}
\section{Experiments}
\label{sec:exp}
We evaluate \covo in three nonlinear robotic systems. 
To demonstrate the effectiveness of \covo, we first compare it with the standard \mppi, in both simulation and the real world.
To further understand the difference between \covo~and~\mppi, we quantify their computational costs and visualize their cost distributions.
Our results show that \covo significantly outperforms \mppi with a more concentrated cost distribution.

\vspace{-10pt}
\subsection{Tasks and Implementations}

\begin{wrapfigure}{r}{0.44\textwidth}
    \centering    
    \vspace{-10pt}\includegraphics[width=0.44\textwidth]{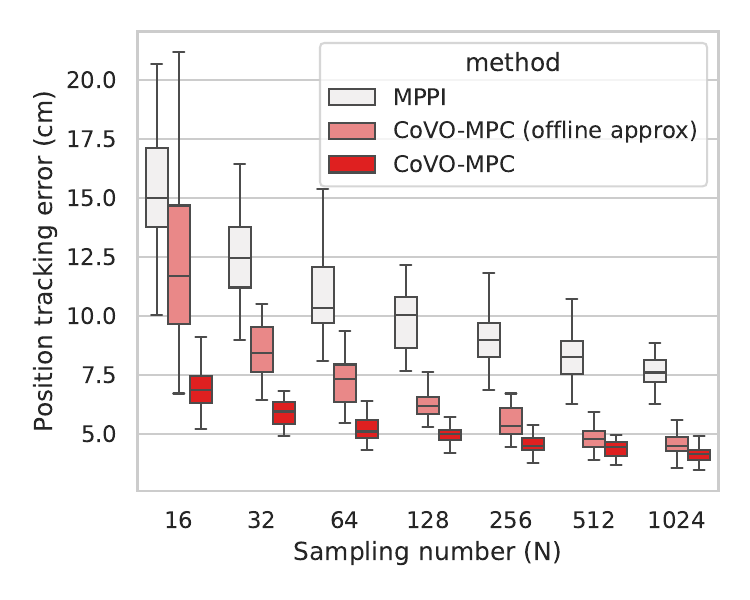}
    \vspace{-35pt}
    \caption{\texttt{Quadrotor} tracking errors as the sample number \(N\) increases.}
    \label{fig:abalation_N}
\end{wrapfigure}

We choose three tasks (illustrated as \fvtest{\cref{fig:env} in \cref{sec:env_detail}}{Fig.3 in \href{https://tinyurl.com/covo-mpc-cmu}{Appendix} B.2}):
(a)~\texttt{CartPole} environment with force applied to the car as control input~\citep{langeReinforcementLearningEnvironments2023}.
(b)~\texttt{Quadrotor} follows zig-zag infeasible trajectories. The action space is the desired thrust and body rate ($\dim(u)=4$).
(c)~\texttt{Quadrotor (real)}: same tasks as (b) but on a real-world quadrotor platform based on the Bitcraze Crazyflie 2.1~\citep{huang2023datt,shi2021neural}. 

For both the simulation and the real quadrotor, all results were evaluated across $3$ different trajectories, each repeatedly executed $10$ times. The implementation details of all tasks can be found in~\fvtest{\cref{sec:env_detail}}{\href{https://tinyurl.com/covo-mpc-cmu}{Appendix} B.2\hide{ \citep{zejiCoVOMPCTheoreticalAnalysis2023}}}. All tasks use the same hyperparameter (\fvtest{\cref{tab:parameters} in \cref{sec:algo_impl}}{Table 3 in \href{https://tinyurl.com/covo-mpc-cmu}{Appendix} B.2\hide{ \citep{zejiCoVOMPCTheoreticalAnalysis2023}}}). 
It is worth noting that~\cref{alg:covo} ensures the covariance matrix's determinant of \covo is identical to that of the \mppi baseline, to keep their sampling volumes the same.

\subsection{Performance and Computational Cost}
\begin{figure}[ht]
    \centering   
    \vspace{5pt}\includegraphics[width=0.9\linewidth]{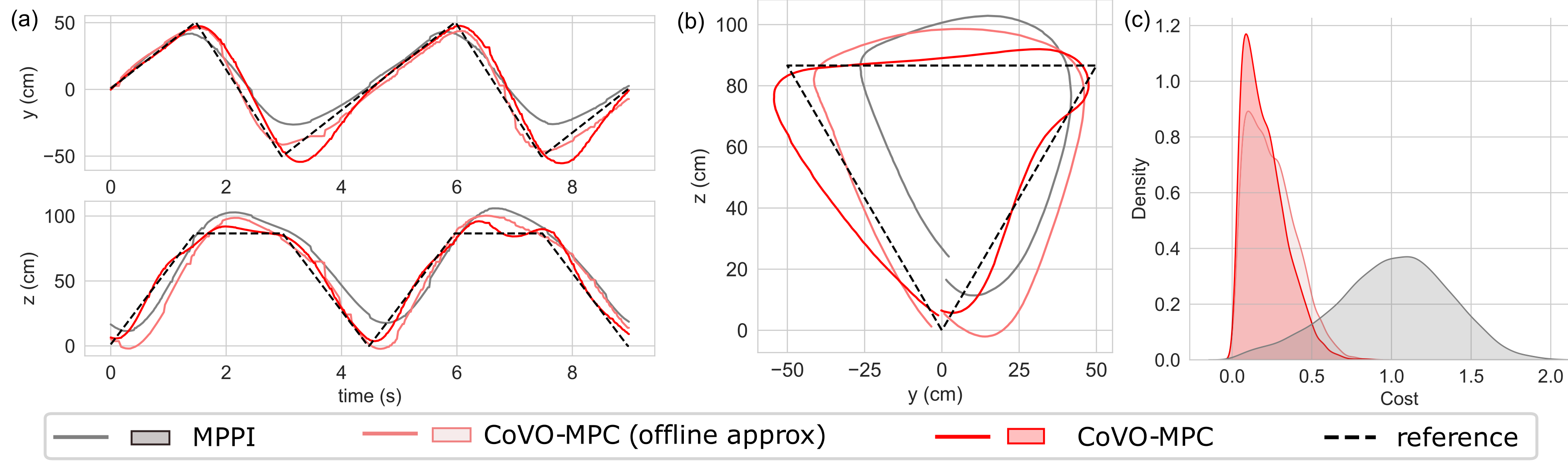}
    \vspace{-10pt}
    \caption{(a-b) Real-world quadrotor trajectory tracking results. \covo can track the challenging infeasible triangle trajectory closer than \mppi. (c) The cost distribution of sampled trajectories at a certain time step in \texttt{Quadrotor} simulation. The cost of \covo is more concentrated and has a lower mean than \mppi.}
    \vspace{-5pt}
    \label{fig:traj_cost}
\end{figure}

\cref{tab:main} illustrates the evaluated cost for each task. \covo outperforms \mppi across all tasks with varying performance gains from \(43\%\) to \(54\%\). 
In simulations, the performance of \covo (offline approx.) stays close to \covo by approximating the Hessian matrix offline using simple nominal PID controllers, which implies the effectiveness of \covo comes from the optimized non-trivial pattern of $\Sigma_t$ (\fvtest{\cref{fig:cov_heatmap} in \cref{sec:dist_vis}}{Fig. 4 in \href{https://tinyurl.com/covo-mpc-cmu}{Appendix} C}) rather than its precise values.
When transferred to the real-world quadrotor control, the offline approximation's performance degrades due to sim-2-real gaps but still outperforms \mppi by a significant margin ($22\%$). 
The real-world tracking results are visualized in~\cref{fig:traj_cost}, where \covo can effectively track the desired triangle trajectory while \mppi fails.
\vspace{8pt}
\begin{table}[ht]
    \centering
    \begin{tabular}{cccc}
       Tasks          & \texttt{CartPole}        & \texttt{Quadrotor}       & \texttt{Quadrotor (real)} \\
        \hline \hline
        \covo                  & $\mathbf{0.70 \pm 0.21}$ & $\mathbf{3.71 \pm 0.37}$ & $\mathbf{8.36 \pm 4.86}$  \\
        \covo (offline approx.) & $0.71 \pm 0.23$          & $3.86 \pm 0.34$          & $12.09 \pm 6.33$          \\
        \hline
        \mppi                  & $1.52 \pm 0.46$          & $6.48 \pm 0.64$          & $15.49 \pm 5.82$          \\
    \end{tabular}
    \caption{The cost associated with \covo compared with that of \mppi while tracking an infeasible zig-zag trajectory. For \texttt{Quadrotor} and \texttt{Quadrotor (real)}, the value indicates tracking error in centimeters. }
    \label{tab:main} 
\end{table}

Besides optimality, another important aspect of sampling-based MPC is its sampling efficiency. To understand \covo's sampling efficiency, we evaluate all methods in quadrotor simulation with various sampling numbers $N$. The results in \cref{fig:abalation_N} show that \covo and its approximated variant with fewer samples outperforms the standard \mppi algorithm with much more samples. 

To further understand the performance difference between \covo and \mppi, we plot the cost distributions of their sampled trajectories at a particular time step in~\cref{fig:traj_cost}(c). 
The cost of \covo is more concentrated and has a lower mean than \mppi, which directly implies the effectiveness of the optimal covariance procedure of \covo. In other words, \covo can automatically adapt to different optimization landscapes while \mppi cannot.

\begin{wraptable}{r}{0.65\textwidth}
    \centering
    \begin{tabular}{m{0.16\textwidth}cc}
        \textbf{Algorithm}     & \textbf{Online Time (ms)} & \textbf{Offline Time (ms)} \\ \hline\hline
        \covo                  & $9.22 \pm 0.39$           & $0.26 \pm 0.47$            \\
        Offline approx. & $2.03 \pm 0.53$           & $2179.59 \pm 4.57$         \\
        \hline
        \mppi                  & $2.17 \pm 0.67$           & $2.16 \pm 0.29$            \\
    \end{tabular}
    \caption{Computational time comparison.}
    \label{tab:comp}
\end{wraptable}

As expected, the superior performance of \covo comes with extra computational costs for the Hessian matrix $D_t$ and the optimal covariance $\Sigma_t$. Therefore, we quantify extra computational burdens in \cref{tab:comp}. 
While \covo indeed requires more computation, the offline approximation of \covo has the same computational cost as \mppi but with better performance. 
\vspace{-10pt}

\section{Limitations and Future Work}
While \covo~introduces new theoretical perspectives and shows compelling empirical performance, it currently relies on the differentiability of the system. Our future efforts will focus on adapting our theory and algorithm to more general scenarios. 
We also aim to investigate the finite-sample analysis of sampling-based MPC with a focus on the variance of the algorithm output. Generalizing our results to receding-horizon settings~\citep{lin2021perturbation,yu2020power} is also interesting. Moreover, \covo~holds substantial promise for MBRL, so we plan to integrate \covo~with the MBRL framework, leveraging learned dynamics, value functions, or policies. We also recognize there exist efficient online approximations of \covo, e.g., reusing previous Hessians or covariances, which could lead to more efficient implementations.

\bibliography{ref}

\fvtest{\appendix\label{sec:appendix}\section{Proof Details}
\subsection{Proof Details for Quadratic \texorpdfstring{$J(U)$}{J(U)}}
\label{apd:Quad_J}
We first introduce Slutsky's Theorem below. 

\begin{lemma}[Slutsky' theorem]
    Let $X_n, Y_n$ be sequences of scalar/vector/matrix random variables. If $X_n$ converges in probability to a random element $X$ and $Y_n$ converges in probability to a constant $c$, then$X_n+Y_n \stackrel{p}{\rightarrow} X+c$;$X_n Y_n \stackrel{}{\rightarrow} X c$; $X_n / Y_n \stackrel{p}{\rightarrow} X / c$, provided that $c$ is invertible, where $\stackrel{p}{\rightarrow}$ denotes convergence in probability.
\end{lemma}

With Slutsky's Theorem, we now give the detailed proof of \cref{theo:1_contraction}.

\bigskip

\begin{proof}
Firstly, we focus on the case \( U_{\mathrm{in}} = U^* \) and show the optimal point \( U^* \) serves as a fixed point of \eqref{eq:MPPI}. Then, we move on to the general case. 

\noindent\textbf{The case \( U_{\mathrm{in}} = U^* \).} The quadratic nature of the function $J(U)$ allows us to express it as \[ U^\top D U + U^\top d + J_c = J(U^*) + (U-U^*)^\top D (U-U^*) ,\] with \( J_c \) being a constant and \( J^* \) defined as \( J(U^*) \). To show the convergence of \eqref{eq:MPPI}, we separately analyze the distribution and convergence properties of the numerator and denominator of \eqref{eq:MPPI}. 

Our first step is to demonstrate that the denominator converges to a constant in probability. To this end, we calculate the expectation of \( \exp(-\frac{1}{\lambda}J(\mathcal{U})) \). 

\begin{align*}
&    \mathbb{E}(\exp(-\frac{1}{\lambda}J(\mathcal{U}))) \nonumber 
\\
&= \int \frac{\exp(-\frac{J^*}{\lambda})}{\sqrt{(2\pi)^k}\det^{\frac{1}{2}}{|\Sigma}|}\exp(-\frac{1}{\lambda}(U-U^*)^\top D (U-U^*) -\frac{1}{2} (U-U^*)^\top \Sigma^{-1} (U-U^*)) dU \nonumber
 \\
 &= \frac{\exp(-\frac{J^*}{\lambda})}{\det^{\frac{1}{2}} |{I+\frac{2}{\lambda}D\Sigma}|}.\nonumber
\end{align*}
Utilizing the weak law of large numbers, we observe that \( \frac{1}{N}\sum_i  \exp(-\frac{J(U_i)}{\lambda}) \) converges in probability to \( \mathbb{E}(\exp(-\frac{J(\mathcal{U})}{\lambda})) = \frac{\exp(-\frac{J^*}{\lambda})}{\det^{\frac{1}{2}} |{I+\frac{2}{\lambda}D\Sigma}|} \). This convergence is due to the \( U_i \)'s being independent and identically distributed (i.i.d.) samples from the random variable \( \mathcal{U} \). Notably, convergence in probability implies convergence in distribution; hence,  \( \frac{1}{N}\sum_i  \exp(-\frac{J(U_i)}{\lambda}) \) also converges in distribution to \( \frac{\exp(-\frac{J^*}{\lambda})}{\det^{\frac{1}{2}} |{I+\frac{2}{\lambda}D\Sigma}|} \). 

Next, we focus on analyzing the convergence properties of the numerator by first computing the expectation of \( \mathcal{U} \exp(-\frac{J(\mathcal{U})}{\lambda}) \):

\begin{align}
&\mathbb{E}(\mathcal{U}\exp(-\frac{1}{\lambda}J(\mathcal{U}))) \nonumber\\ 
&= \int \frac{U\exp(-\frac{J^*}{\lambda})}{\sqrt{(2\pi)^k}\det^{\frac{1}{2}}{|\Sigma}|}\exp(-\frac{1}{\lambda}(U-U^*)^\top D (U-U^*) -\frac{1}{2} (U-U^*)^\top \Sigma^{-1} (U-U^*))dU \nonumber\\
&= \int \frac{U\exp(-\frac{J^*}{\lambda})}{\sqrt{(2\pi)^k}\det^{\frac{1}{2}}{\Sigma}} \exp( -\frac{1}{2} (U-U^*)^\top (\Sigma^{-1}+\frac{2}{\lambda}D) (U-U^*)) dU\nonumber\\
&= \frac{U^*\exp(-\frac{J^*}{\lambda})}{\det^{\frac{1}{2}} {(I+\frac{2}{\lambda}D\Sigma})} \nonumber
\end{align}

Combining the results from the denominator and numerator, using Slutsky's Theorem, we are able to show that $U_{\mathrm{out}}\xrightarrow[p]{} U^*$. 

We next show the general contraction property of MPPI where $U_{\mathrm{in}}$ may not necessarily be $U^*$. 

\bigskip

\noindent\textbf{General case. }Recall that now $\mathcal{U} \sim \mathcal{N}(U_{\mathrm{in}},\Sigma)$. And the cost function can be reorganized into 
\begin{align}
    J(U)& = (U-U^*)^\top D (U-U^*) + J^*  \nonumber \\
     & = (U-U_{\mathrm{in}})^\top D (U-U_{\mathrm{in}}) + (U_{\mathrm{in}}-U^*)^\top D (U_{\mathrm{in}}-U^*) \nonumber \\
     & +(U_{\mathrm{in}}-U^*)^\top D (U-U_{\mathrm{in}}) + (U-U_{\mathrm{in}})^\top D (U_{\mathrm{in}}-U^*) + J^*\nonumber
\end{align}

We then again calculate the expectation of the denominator: 
\begin{align}
      &\mathbb{E}(\exp(-\frac{1}{\lambda}J(\mathcal{U}))) \nonumber \\
      & = \int \frac{\exp(-\frac{J(U)}{\lambda})}{\sqrt{(2\pi)^k}\det^{\frac{1}{2}}{\Sigma}}\exp( -\frac{1}{2} (U-U_{\mathrm{in}})^\top \Sigma^{-1} (U-U_{\mathrm{in}})) dU \nonumber \\
     &= \int \frac{\exp(-\frac{J^* + (U_{\mathrm{in}}-U^*)^\top D (U_{\mathrm{in}}-U^*)}{\lambda})}{\sqrt{(2\pi)^k}\det^{\frac{1}{2}}{\Sigma}} \cdot \exp(-\frac{1}{\lambda}(U-U_{\mathrm{in}})^\top D (U-U_{\mathrm{in}}) -\frac{1}{2} (U-U_{\mathrm{in}})^\top \Sigma^{-1} (U-U_{\mathrm{in}})) \cdot \nonumber \\
     & \exp(-\frac{1}{\lambda}(U_{\mathrm{in}}-U^*)^\top D (U-U_{\mathrm{in}}) + (U-U_{\mathrm{in}})^\top D (U_{\mathrm{in}}-U^*)) dU \nonumber \\
& =C \int  \exp(-\frac{1}{2}(U+(\frac{2}{\lambda}\Sigma^{-1} + D)^{-1} D (U_{\mathrm{in}}-U^*))^\top (\Sigma^{-1} + \frac{2}{\lambda} D) (U+(\frac{\lambda}{2}\Sigma^{-1} +  D)^{-1} D (U_{\mathrm{in}}-U^*)))dU \nonumber
\end{align}
Notice that the last equality coming from a change of variable that we substitute $U$ with $U+U_{\mathrm{in}}$, 
and $C$ is defined as 
$$
    C = \frac{\exp(-\frac{J^* + (U_{\mathrm{in}}-U^*)^\top D (U_{\mathrm{in}}-U^*)}{\lambda}+\frac{2}{\lambda^2}(U_{\mathrm{in}}-U^*)^\top D (\Sigma^{-1}+\frac{2}{\lambda}D)^{-1}D (U_{\mathrm{in}}-U^*) )}{\det^{\frac{1}{2}}{\Sigma}\sqrt{(2\pi)^k}}.
$$ So the expectation of the denominator is 
$
  \mathbb{E}(\exp(-\frac{1}{\lambda}J(\mathcal{U}))) = C \frac{\sqrt{(2 \pi)^k} \det ^{\frac{1}{2}}\Sigma}{\det^{\frac{1}{2}} {I+\frac{2}{\lambda}D\Sigma}}
$. 

Similarly, for the numerator, we have
\begin{align}
&\mathbb{E}(\mathcal{U}\exp(-\frac{1}{\lambda}J(\mathcal{U}))) \nonumber \\ 
&=C \int (U+U_{\mathrm{in}})\exp(-\frac{(U+(\frac{2}{\lambda}\Sigma^{-1} + D)^{-1} D (U_{\mathrm{in}}-U^*))^\top (\Sigma^{-1} + \frac{2}{\lambda} D) (U+(\frac{\lambda}{2}\Sigma^{-1} +  D)^{-1} D (U_{\mathrm{in}}-U^*)))}{2}dU \nonumber \\
&= U_{\mathrm{in}} \mathbb{E}(\exp(-\frac{1}{\lambda}J(\mathcal{U}))) + \frac{2}{\lambda}(\Sigma^{-1} + \frac{2}{\lambda} D)^{-1} D (U^*-U_{\mathrm{in}}) \mathbb{E}(\exp(-\frac{1}{\lambda}J(\mathcal{U})))\nonumber
\end{align}
With the weak law of large number $ \frac{1}{N}\sum_i \exp(-\frac{1}{\lambda}J(U_i)) \xrightarrow[p]{} \mathbb{E}(\exp(-\frac{1}{\lambda}J(\mathcal{U})))$ and$ \frac{1}{N}\sum_i U_i \exp(-\frac{1}{\lambda}J(U_i)) \xrightarrow[p]{} \mathbb{E}(\mathcal{U}\exp(-\frac{1}{\lambda}J(\mathcal{U})))$. We then apply Slutsky's theorem here and can get 
$$
U_{\mathrm{out}} = \frac{\frac{1}{N}\sum U\exp(-\frac{1}{\lambda}J(U))}{\frac{1}{N}\sum\exp(-\frac{1}{\lambda}J(U))}  \xrightarrow[p]{} U_{\mathrm{in}}  + \frac{2}{\lambda}(\Sigma^{-1} + \frac{2}{\lambda} D)^{-1} D (U^*-U_{\mathrm{in}}) . 
$$
Therefore, we have 
\begin{equation}
(U_{\mathrm{out}} - U^*) \xrightarrow[p]{}  (I-\frac{2}{\lambda}(\Sigma^{-1}+\frac{2}{\lambda}D)^{-1}D)(U_{\mathrm{in}} - U^*)  =  (\frac{2}{\lambda} \Sigma D+I)^{-1} (U_{\mathrm{in}} - U^*) .
\label{eq:quad_conv_inp}
\end{equation}
Because the RHS of \eqref{eq:quad_conv_inp} is a constant, we are allowed to take a norm on both sides to get $\left\|U_{\mathrm{out}} - U^*\right\| \xrightarrow[p]{}  \left\|(U_{\mathrm{in}} - U^*) (\frac{2}{\lambda} \Sigma D+I)^{-1}\right\|$.
Then, with Cauchy-Schwartz inequality, we can have $$\left\| (\frac{2}{\lambda} \Sigma D+I)^{-1} (U_{\mathrm{in}} - U^*)\right\|_2 \leq \left\|U_{\mathrm{in}} - U^* \right\|_2 \left\|(\frac{2}{\lambda} \Sigma D+I)^{-1}\right\|_2.$$
This leads to $$\frac{\left\|U_{\mathrm{out}} - U^*\right\|_2}{\left\|U_{\mathrm{in}} - U^* \right\|_2 } \xrightarrow[p]{}  \mathrm{Contraction Rate}\leq \left\|(\frac{2}{\lambda} \Sigma D+I)^{-1}\right\|_2 \leq 1$$

By Continuous Mapping Theorem, $J(\mathcal{U})$ is a continuous function of $\mathcal{U}$, so
\begin{align}
J(U_{\mathrm{out}}) &\xrightarrow[p]{}   (U_{\mathrm{in}} - U^*)^\top  (\frac{2}{\lambda} \Sigma D+I)^{-\top} D (\frac{2}{\lambda} \Sigma D+I)^{-1} (U_{\mathrm{in}} - U^*) + J^* \nonumber \\ 
&= (U_{\mathrm{in}} - U^*)^\top D^{\frac{1}{2}} D^{-\frac{1}{2}}  (\frac{2}{\lambda} \Sigma D+I)^{-\top} D^{\frac{1}{2}} D^{\frac{1}{2}} (\frac{2}{\lambda} \Sigma D+I)^{-1}  D^{-\frac{1}{2}} D^{\frac{1}{2}} (U_{\mathrm{in}} - U^*) +J^* \nonumber\\
&= (D^{\frac{1}{2}} (U_{\mathrm{in}} - U^*))^\top (\frac{2}{\lambda} D^{\frac{1}{2}} \Sigma D^{\frac{1}{2}}+I)^{-\top} (\frac{2}{\lambda} D^{\frac{1}{2}} \Sigma D^{\frac{1}{2}}+I)^{-1}(D^{\frac{1}{2}} (U_{\mathrm{in}} - U^*))+ J^*\nonumber\\
&= \left\|   (\frac{2}{\lambda} D^{\frac{1}{2}} \Sigma D^{\frac{1}{2}}+I)^{-1}(D^{\frac{1}{2}} (U_{\mathrm{in}} - U^*))\right\|_2^2 +J^*\nonumber \\
& \leq \left\|   (\frac{2}{\lambda} D^{\frac{1}{2}} \Sigma D^{\frac{1}{2}}+I)^{-1} 
\right\|_2^2
\left\| (D^{\frac{1}{2}} (U_{\mathrm{in}} - U^*))\right\|_2^2 +J^* \nonumber \\
& = \left\|   (\frac{2}{\lambda} D^{\frac{1}{2}} \Sigma D^{\frac{1}{2}}+I)^{-1} 
\right\|_2^2 (J(U_{\mathrm{in}})-J^*) +J^*
\end{align}
where the inequality comes from Cauchy-Schwartz inequality and the last equation comes from $J(U_{\mathrm{in}})-J^* = (U_{\mathrm{in}} - U^*)^\top D (U_{\mathrm{in}} - U^*) = \left\| (D^{\frac{1}{2}} (U_{\mathrm{in}} - U^*))\right\|_2^2 $. As a result, the above leads to
$$
        J(U_{\mathrm{out}}) - J^*\xrightarrow[p]{} J_{\mathrm{out}} - J^*\leq (J(U_{\mathrm{in}})-J^*)\left\|(I+\frac{2}{\lambda}D^{\frac{1}{2}}\Sigma D^{\frac{1}{2}})^{-2} \right\|.
        $$
\end{proof}
The contraction of the cost function have a single-step contraction rate of $\left\|(I+\frac{2}{\lambda}D^{\frac{1}{2}}\Sigma D^{\frac{1}{2}})^{-2} \right\| $.

This means that if we apply \eqref{eq:MPPI} to solve \eqref{eq:cost_func} iteratively with infinite samples,  then it takes $O(\log \frac{1}{\epsilon})$ steps to reach $J(U^{\mathrm{out}})-J^*\leq \epsilon$.
We also notice that the contraction shares the same form with regularized newtown's method. Regularized Newtown's method has the following form 
$$
    U_{\mathrm{in}}^{newton}=U_{\mathrm{in}}-\left(\nabla^2 J\left(U_{\mathrm{in}}\right)+C I\right)^{-1} \nabla f\left(U_{\mathrm{in}}\right) = U_{\mathrm{in}}-\left( D +C I\right)^{-1} D (U_{\mathrm{in}} -U^*) 
$$

Organizing the RHS, we can find $U_{\mathrm{in}}^{newton} = (\frac{D}{C}+I)^{-1}(U_{\mathrm{in}} -U^*)$. 
When taking $\Sigma= \frac{\lambda}{2C} I$ (Identity is a common choice for MPPI), the expectation of the output control as in \cref{eq:expectation_contraction} follows exactly the regularized Newtown's method, which runs Newtown's method on regularized cost function $J(U)+ \lambda (U-U_{\mathrm{in}})^\top (U-U_{\mathrm{in}})$.
Such regularized Newton's method, in general, has contraction, which is consistent with the \cref{theo:1_contraction} above. Further, as 
Theorem 2 of  \citet{mishchenkoRegularizedNewtonMethod2023} shows, with properly (sometimes adaptively) chosen $\lambda$ and $\Sigma$ (for instance, $\Sigma = I,\lambda \propto \sqrt{\left\|D (U_{\mathrm{in}}-U^*)\right\|}$), we only need $O(\log\log(\frac{1}{\epsilon}))$ iterations for MPPI to converge.\footnote{The convergence rate only holds locally as stated in the literature. However, this neighborhood is actually defined by the inverse of Hessian's Lipschitz. For quadratic function, the Hessian is a constant. Therefore, it holds globally.} 

\subsection{Proof Details for Optimal Covariance Design }
\label{apd:opt_cov}
Initially, let's assume that $\Sigma$ and $D$ are commutative. The justification for this assumption will be elaborated upon in the proof of \cref{coro:sigma}. Under this assumption, $\Sigma$ and $D$ share the same set of eigenvectors, denoted by the matrix $V$. Correspondingly, we define $\Lambda$ and $\mathcal{O}$, both in $\mathbb{R}^{mH \times mH}$, where $\Lambda = \text{diag}([\Lambda]_1, [\Lambda]_2, \ldots, [\Lambda]_{mH})$ and $\mathcal{O} = \text{diag}([\mathcal{O}]_1, [\mathcal{O}]_2, \ldots, [\mathcal{O}]_{mH})$.

Moreover, let $\Lambda_{\mathrm{in}} \in \mathbb{R}^{mH}$ represent the vector formed by the coordinates $U_{\mathrm{in}} - U^*$ in the basis of $V$. In mathematical terms, this is expressed as $ \Lambda_{\mathrm{in}} = V (U_{\mathrm{in}} - U^*) = [[\Lambda_{\mathrm{in}}]_1, [\Lambda_{\mathrm{in}}]_2, \ldots, [\Lambda_{\mathrm{in}}]_{mH}]^\top$. With the above setup, here we give the detailed proof of \cref{theo:optimal_sigma}.
\begin{proof}
The objective function is
$$
 (U_{\mathrm{in}} - U^*)^\top  (\frac{2}{\lambda} \Sigma D+I)^{-\top} D(\frac{2}{\lambda} \Sigma D+I)^{-1} (U_{\mathrm{in}} - U^*)
$$
And substitute  $\Sigma $ with $ V^\top \Lambda V$ and   $D $ with $ V^\top \mathcal{O} V$. The optimization objective is equivalent to $$(U_{\mathrm{in}} - U^*)^\top  (\frac{2}{\lambda} V^\top \Lambda V  V^\top \mathcal{O} V + I)^{-\top} V^\top \mathcal{O} V (\frac{2}{\lambda} V^\top \Lambda V V^\top \mathcal{O} V+I)^{-1} (U_{\mathrm{in}} - U^*),$$ 
which simplifies to $(U_{\mathrm{in}} - U^*)^\top V^\top (\frac{2}{\lambda}  \Lambda \mathcal{O} + I)^{-1}  \mathcal{O}  (\frac{2}{\lambda}  \Lambda  \mathcal{O} +I)^{-1} V(U_{\mathrm{in}} - U^*)$. Therefore, the objective of the optimization problem is:

$$
   [V(U_{\mathrm{in}} - U^*)]^\top (\frac{2}{\lambda}  \Lambda \mathcal{O} + I)^{-1}  \mathcal{O}  (\frac{2}{\lambda}  \Lambda  \mathcal{O} +I)^{-1} [V(U_{\mathrm{in}} - U^*)] $$
Because $\Lambda, \mathcal{O}$ are now diagonal matrix, we can rewrite it into:
    $$
     \sum_{i=1}^{mH} [\Lambda_{\mathrm{in}}]_i^2  \frac{[\mathcal{O}]_i}{(\frac{2}{\lambda}[\mathcal{O}]_i [\Lambda]_i+1)^2}
$$
The constraint optimization problem can be rewrite into:
\begin{align*}
    \min_{[\Lambda]} & \sum_{i=1}^{mH} [\Lambda_{\mathrm{in}}]_i^2  \frac{[\mathcal{O}]_i}{(\frac{2}{\lambda}[\mathcal{O}]_i [\Lambda]_i+1)^2} \\
    s.t. & \prod [\Lambda]_i \leq \alpha, [\Lambda]_i>0
\end{align*}
Notice that the objective is monotone with respect to any $[\Lambda]_i$. We now  employ the Lagrangian multiplier method here.

The Lagrangian \( \mathcal{L} \) for the optimization problem combines the objective function with the constraint using a Lagrangian multiplier \( \mu \). The problem is:
   \begin{align*}
   \min_{[\Lambda]} & \sum_{i=1}^{mH} [\Lambda_{\mathrm{in}}]_i^2  \frac{[\mathcal{O}]_i}{(\frac{2}{\lambda}[\mathcal{O}]_i [\Lambda]_i+1)^2} \\
   \text{s.t.} & \prod [\Lambda]_i \leq \alpha, [\Lambda]_i>0
   \end{align*}
   The corresponding Lagrangian is:
   \begin{equation*}
   \mathcal{L}([\Lambda], \mu) = \sum_{i=1}^{mH} [\Lambda_{\mathrm{in}}]_i^2  \frac{[\mathcal{O}]_i}{(\frac{2}{\lambda}[\mathcal{O}]_i [\Lambda]_i+1)^2} + \mu \left(\prod_{i=1}^{mH} [\Lambda]_i - \alpha\right)
   \end{equation*}
To find the minimum, we set the derivative of \( \mathcal{L} \) with respect to \( [\Lambda]_i \) and \( \mu \) to zero. 

   For each \( i = 1, \cdots, mH \), the derivative with respect to \( [\Lambda]_i \) is:
   \begin{equation*}
   \frac{\partial L}{\partial [\Lambda]_i} = -2 [\Lambda_{\mathrm{in}}]_i^2  \frac{[\mathcal{O}]_i^2}{(\frac{2}{\lambda}[\mathcal{O}]_i [\Lambda]_i+1)^3} + \mu \frac{\partial}{\partial [\Lambda]_i} \left(\prod_{j=1}^{mH} [\Lambda]_j\right) = 0
   \end{equation*}

   The derivative of the product term \( \prod_{j=1}^{mH} [\Lambda]_j \) with respect to \( [\Lambda]_i \) is the product of all \( [\Lambda]_j \) terms except for \( [\Lambda]_i \). 

\begin{equation}
   \frac{[\mathcal{O}]_{i}^2 [\Lambda_{\mathrm{in}}]_i^2 [\Lambda]_{i}}{(1+\frac{2}{\lambda}[\Lambda]_{i} [\mathcal{O}]_{i})^3} = \mu \prod_{j=1}^{mH} [\Lambda]_j, \forall i=1,\cdots,mH.
   \label{eq:lag_mul}
   \end{equation}

It is clear that $\prod_{j=1}^{mH} [\Lambda]_j$ is a constant for any $i$. According to the Lagrange multiplier theorem any maximum or minimum corresponding to a $\mu$ that solves \eqref{eq:lag_mul}. Along with the fact that the problem is monotonically decreasing with respect to any $[\Lambda]_i$, and the biggest allowed $\prod_{j=1}^{mH} [\Lambda]_j$ is $\alpha$. Therefore,  the solution is:
\begin{align*}
   \frac{[\mathcal{O}]_{i}^2 [\Lambda_{\mathrm{in}}]_i^2 [\Lambda]_{i}}{(1+\frac{2}{\lambda}[\Lambda]_{i} [\mathcal{O}]_{i})^3} &= Const, \forall i=1,\cdots,mH. \\
   s.t. & \prod_{j=1}^{mH} [\Lambda]_j = \alpha
\end{align*}

\end{proof}
\begin{remark}
Without further knowledge of the gap between $U^*$ and $U_\mathrm{in}$, the most reasonable way is to assume that  $U_\mathrm{in} - U^* \sim \mathcal{N}(0,I)$. We define $C_J =  (\frac{2}{\lambda} \Sigma D+I)^{-\top} D(\frac{2}{\lambda} \Sigma D+I)^{-1} $ Then $\mathbb{E}(U_\mathrm{in} - U^*)^\top C_J (U_\mathrm{in} - U^*) = \mathrm{Tr}(C_J)$. Moreover, if we fix the norm of   $\left\|(U_\mathrm{in} - U^*)\right\| = 1$. The mini-max problem of $(U_\mathrm{in} - U^*)^\top C_J (U_\mathrm{in} - U^*)$ with respect to $C_J$ 
\begin{align*}
    \min_{C_J}\max_{U^*}& (U_\mathrm{in} - U^*)^\top C_J (U_\mathrm{in} - U^*) \\
    s.t.& \left\|(U_\mathrm{in} - U^*)\right\| = 1
\end{align*}
has a solution of $C_J \propto I$.
\label{rem:apd_opt_cov}
\end{remark}

Now we give the proof of \cref{coro:sigma}.
\begin{proof}
From the remark, our goal is to design 
    $(\frac{2}{\lambda} \Sigma D+I)^{-\top} D(\frac{2}{\lambda} \Sigma D+I)^{-1} \propto I $ when $\lambda \rightarrow 0$. And  $(\frac{2}{\lambda} \Sigma D+I)^{-\top} D(\frac{2}{\lambda} \Sigma D+I)^{-1}  = (\frac{2}{\lambda} \Sigma D^{1/2}+D^{-1/2})^{-\top} (\frac{2}{\lambda} \Sigma  D^{1/2}+D^{-1/2})^{-1} $. 
    Therefore $$(\frac{2}{\lambda} \Sigma D^{1/2}+D^{-1/2})^{\top} (\frac{2}{\lambda} \Sigma  D^{1/2}+D^{-1/2}) \propto I$$. When $\lambda \rightarrow 0$. $(\frac{2}{\lambda} \Sigma D^{1/2}+D^{-1/2})^{\top} (\frac{2}{\lambda} \Sigma  D^{1/2}+D^{-1/2}) \approx \frac{4}{\lambda^2}D^{\frac{1}{2}}\Sigma^2 D^\frac{1}{2} \propto I$. So, we can say that $\Sigma \propto D^{-\frac{1}{2}}$. And $\Sigma = (\alpha \det D^{1/2})^{\frac{1}{mH}} D^{-1/2}$

\end{proof}

\subsection{Proof Details for Non-Quadratic \texorpdfstring{$J(U)$}{J(U)}}

\subsubsection{Strongly Convex}
\label{apd:proof_str_conv}
In numerous practical scenarios, the total cost function $J$ exhibits strong convexity without being quadratic. As an example, this is often encountered in RL reward designs, where instead of the standard quadratic form $x^\top Q x$, a nonlinear convex function $Q(x)$ is used. Motivated by such cases, we here consider the convergence property of a $\beta$-strongly convex total cost function $J(U)$ that satisfies the inequality $J(U') \geq J(U)+\nabla J(U)^\top(U'-U)+\frac{\beta}{2}\|U'-U\|^2$ for all $U'$ and $U$.
\begin{statement}{Theorem 4}
    Given a $\beta$-strongly convex function $J(U)$ which has Lipschitz continuous derivatives, that is $\frac{\partial J}{\partial U} $ is $ L_d$-Lipschitz,  $U_{\mathrm{out}}$ converges to a neighborhood of the optimal point $U^*$ in probability as the number of samples $N$ tends to infinity, i.e., $U_{\mathrm{out}} \xrightarrow[p]{} U_{c}$. Here $U_c$ satisfies $\left\|U_{c} - U^*\right\|   \leq  \frac{\lambda}{\beta}\frac{2\left\|  \Sigma^{-1} \right\|}{(1+\frac{\lambda}{\beta \left\|\Sigma\right\|})}\left\| U_{\mathrm{in}}-U^{*}\right\| + \left\| U_{\mathrm{error}} \right\|$, where $\left\|U_{\mathrm{error}}\right\| \sim O(\frac{\sqrt{\lambda}}{\beta})$.
    \label{thm:small_lambda_nonlinear}
\end{statement}

In \cref{theo:small_lambda}, notice that $U_{\mathrm{out}}$ consists of two parts: a contraction from $U_{\mathrm{in}}$ to $U^*$ of the form $O(\frac{\lambda}{\beta}\left\|U_{\mathrm{in}}-U^*\right\|)$ and an error term (coming from the nature of the softmax style zeroth order method) $O(\frac{\sqrt{\lambda}}{\beta})$. 
 This implies that with a small $\lambda$, MPPI maintains a linear contraction rate of $O(\frac{\lambda}{\beta})$, which approaches zero as $\lambda \ll \beta$. The remaining residual error $U_{\mathrm{error}}$, of $O(\frac{\sqrt{\lambda}}{\beta})$, suggests the eventual convergence of MPPI to a neighborhood around the optimal point of size $O(\frac{\sqrt{\lambda}}{\beta})$. Notably, this contraction rate differs from that in \cref{sec:quad_conv} because 
non quadratic $J(U)$ leads to slightly larger constant in some inequalities in the proof.
We now provide a proof for \cref{theo:small_lambda}.
\begin{proof}
According to Importance Sampling, a random variable $\mathcal{U}$'s expectation under distribution $P(U)$ can be estimated by samples from distribution $Q(U)$ with the following equation:
\begin{equation}
    \hat{\mathbb{E}}_{P} \mathcal{U} = \frac{1}{N} \sum_{i=0}^N U_i \frac{P(U_i)}{Q(U_i)}
    \label{eq:IS}
\end{equation}
Let $$P(U)\propto \mathcal{N}(U\mid U_{\mathrm{in}},\Sigma)\exp(-\frac{J(U)}{\lambda})=\exp(-\frac{J'(U)}{\lambda})$$
where $J'(U)$ is defined as $J'(U) = U^\top (D+\frac{\lambda}{2}\Sigma^{-1})U + U^\top (d-
\lambda \Sigma^{-1} U_{\mathrm{in}})$.
We also define $Q(U) \sim \mathcal{N}(U_\mathrm{in},\Sigma)$ and substitute into \eqref{eq:IS}, we can find \eqref{eq:MPPI} is estimating distribution $P(U)\propto \exp(-\frac{J'(U)}{\lambda})$ with samples from distribution $\mathcal{N}(U_\mathrm{in},\Sigma)$. Therefore, from weak law of large number, we have $U_\mathrm{out}\xrightarrow[p]{} \mathbb{E}_{P} \mathcal{U}$. This means that to prove \cref{theo:small_lambda}, we only need to bound the distance between the distribution's mean $\mathbb{E}_{P} \mathcal{U}$ and $U^*$, which can be bounded by a combination of two parts. The first part is the gap between $\mathbb{E}_{P} \mathcal{U}$, and the optimal point of $J'(U)$; the second part is the gap between the optimal point of $J'(U)$ and $U^*$. We bound these two parts separately below.

\noindent\textbf{Part 1: Gap between $\mathbb{E}_{P} U$ and the optimal point of $J'(U)$.}  
The approach in this section begins by transforming the function \( J'(U) \) into \( \Tilde{J}(U) = J'(U - U^{*'}) \), where \( U^{*'} \) represents the optimizer of \( J' \).  We denote $\Tilde{p}(U)$ as the distribution determined by $\Tilde{J}(U)$ that $\Tilde{p}(U) \propto \exp(-\Tilde{J}(U)) = \exp(-J'(U-U^{*'}))$. Therefore, $\mathbb{E}_{P}(\mathcal{U}) = \mathbb{E}_{\Tilde{P}}(\mathcal{U})+U^{*'}$, and we have

\begin{equation}
\mathbb{E}_{P}\mathcal{U} = \frac{\int U \exp(-J'(U)/\lambda)dU}{\int \exp(-J'(U)/\lambda)dU} = \underbrace{\frac{\int U \exp(-\Tilde{J}(U)/\lambda)dU}{\int \exp(-\Tilde{J}(U)/\lambda)dU}}_{\mathbb{E}_{\Tilde{P}}(\mathcal{U})}+U^{*'} 
\label{eq:epu}
\end{equation}
This leads to a bound on $\mathbb{E}_{P}(\mathcal{U})$ that $\mathbb{E}_{\Tilde{P}}(\mathcal{U})$. Here we will give the  bound of $\mathbb{E}_{\Tilde{P}}(\mathcal{U})$ by  separately determining the upper bound for the numerator and the lower limit for the denominator, setting the stage for the subsequent analysis.

 This redefinition places the optimal point of \( \Tilde{J}(U) \) at \( \Tilde{J}(\boldsymbol{0}) = 0 \). Moreover, \( \Tilde{J}(U) \) also has Lipschitz continuous deriavtive, satisfying the condition \( \left\|\frac{\partial^2 \Tilde{J}(U)}{\partial U^2}\right\| \leq L_H, \forall U  \). Given that $\Tilde{J}(0) = 0$ and $\frac{\partial \Tilde{J}}{\partial U}=0\mid_{U=0}$, we can bound $\Tilde{J}(U)$ by   $\Tilde{J}(U) \leq \frac{1}{2} L_H \left\|U\right\|^2$. Consequently, we can express this bound as \( \frac{\Tilde{J}(U)}{\lambda} \leq \frac{L_H}{2}\frac{\| U \|^2}{\lambda} \).

We have that $\int \exp(-\Tilde{J}(U)/\lambda) dU \geq \int \exp(-\frac{1}{2}L_H\left\|U\right\|^2/\lambda) dU =  c_d \int_0^{\infty} e^{-\frac{L_H}{2\lambda}x^2} d x $, where $c_d$ is a dimension specific constant (e.g. $c_1 = 2$  $c_2 = 2 \pi$ , $c_3 = 4 \pi$). 
Further, the integral is in the family of Gamma function because$
    \int_0^{\infty} e^{-t^m} d t=\frac{1}{m} \int_0^{\infty} s^{1 / m-1} e^{-s} d s=\frac{1}{m} \Gamma(1 / m)=\Gamma(1+1 / m)
$. By changing the variable, we can get $\int_0^{\infty} e^{-t^m/a} d t = \int_0^{\infty} e^{-v^m} a^{1/m} d v = a^{1/m} \int_0^{\infty} e^{-t^m/a} d t$. By taking $m=2$ and $a = \frac{2\lambda}{L_H}$, and we have
\begin{equation}
\label{eq:str_c_dem}
    \int \exp(-\Tilde{J}(U)/\lambda) dU \geq   c_d c_\Gamma (\frac{2\lambda}{L_H})^{\frac{1}{2}} = c_d c_\Gamma (\frac{2}{L_H})^{\frac{1}{2}} \lambda^{\frac{1}{2}},
\end{equation}
where $c_\Gamma = \Gamma(\frac{3}{2}) \approx 0.886$, is a constant. With the fact that $J$ is $\beta$-strongly convex, we now that $J'$ and $\Tilde{J}$ are at least $\beta$-strongly convex. Meanwhile, the optimal point for $\Tilde{J}$ is $U = 0$. Therefore, 
$
 \Tilde{J}(U) \geq  U^\top \frac{\beta}{2} I U
.$
We can then bound $\int U \exp(-\Tilde{J}(U)/\lambda)dU$ with
\begin{equation}
\int U \exp(-\Tilde{J}(U)/\lambda)dU \leq \int \left\|U\right\| \exp(-\frac{\beta}{2 \lambda} U^\top U)dU = c_d \frac{\lambda}{\beta}
\label{eq:str_c_nom}
\end{equation}
Putting \eqref{eq:str_c_dem} and \eqref{eq:str_c_nom} together, we can bound the gap between the mean $\mathbb{E}\mathcal{U}$ of distribution $P(U)\propto \exp(-\frac{J'(U)}{\lambda})$ with the optimal point $U^{*'} $of $J'$ as follows: 
\begin{equation}
   \left\|  \mathbb{E}_{p} \mathcal{U} - U^{*'} \right\| \leq \frac{\lambda}{\beta} \frac{\sqrt{L_H}}{\sqrt{2\lambda}c_\Gamma} \sim O(\frac{\sqrt{\lambda}}{\beta}).
\end{equation}

\noindent\textbf{Part 2: gap between $U^*$ and optimal point of $J'(U)$}. Recall that 
        $$
            J'(U) = J(U) - \lambda U_{\mathrm{in}}^\top\Sigma^{-1} U + \frac{1}{2} \lambda U^\top \Sigma^{-1} U.
        $$
From the definition of $U^{*'}$, we have
        $
            J'(U^{*'}) \leq J'(U^*)
        $, which leads to
        $$
            J(U^{*'}) - \lambda U_{\mathrm{in}}^\top\Sigma^{-1} U^{*'} + \frac{1}{2} \lambda U^{*'\top }\Sigma^{-1} U^{*'} \leq
            J(U^{*}) - \lambda U_{\mathrm{in}}^\top\Sigma^{-1} U^{*} + \frac{1}{2} \lambda U^{*\top }\Sigma^{-1} U^{*}.
        $$
         Because $J(U)$ is $\beta$-Strongly Convex and $\nabla J(U^*) = 0 $, we have,
        $$
            J(U^{*'}) \geq J(U^*) + (U^{*'}-U^{*})^\top \frac{\beta}{2} I (U^{*'}-U^{*}).
        $$
        As a result, we get
        $$
            J(U^*) + (U^{*'}-U^{*})^\top \frac{\beta}{2} I (U^{*'}-U^{*}) - \lambda U_{\mathrm{in}}^\top\Sigma^{-1} U^{*'} + \frac{1}{2} \lambda U^{*'\top }\Sigma^{-1} U^{*'}\leq J(U^{*}) - \lambda U_{\mathrm{in}}^\top\Sigma^{-1} U^{*} + \frac{1}{2} \lambda U^{*\top }\Sigma^{-1} U^{*}.
        $$
        Denoting $(U^{*'}-U^{*})$ with $\triangle U$, we can get
        $$
            \triangle U^\top \frac{\beta}{2} I \triangle U - \lambda U_{\mathrm{in}}^\top\Sigma^{-1} \triangle U + \frac{1}{2} \lambda \triangle U^\top \Sigma^{-1} \triangle U + \lambda U^{*\top}\Sigma^{-1} \triangle U \leq 0.
        $$
        Simplifying it, we can get
        $$
            \triangle U^\top (\frac{\beta}{2} I +\frac{1}{2} \lambda\Sigma^{-1}) \triangle U  \leq \lambda (U_{\mathrm{in}}-U^{*})^{\top}\Sigma^{-1} \triangle U,
        $$
        which leads to
        $$
            \left\| \triangle U \right\|^2 (\frac{\beta}{2}+\frac{\lambda}{2}\frac{1}{\left\|\Sigma\right\|}) \leq \lambda \left\| \Sigma^{-1} \right\|\left\|U_{\mathrm{in}}-U^{*}\right\|\left\| \triangle U \right\|.
        $$
        We can then get
        $$
            \left\| \triangle U \right\| \leq 2 \lambda \frac{\left\| \Sigma^{-1} \right\|\left\|U_{\mathrm{in}}-U^{*}\right\|}{(\beta+\lambda\frac{1}{\left\|\Sigma\right\|})}
        $$
       
    Putting the two parts together concludes the proof of the theorem.
    \end{proof}

\subsubsection{Linear Systems with Nonlinear Residuals}
\label{apld:proof-res}

Before we formally state our results, we briefly introduce a tool we use in this section, the exponential family. A family $\left\{P_{\boldsymbol{\eta}}\right\}$ of distributions forms an $s$-dimensional exponential family if the distributions $P_{\boldsymbol{\eta}}$ have densities of the form:
$$
P(U ; \eta)=\exp \left[\sum_{i=1}^s \eta_i T_i(U)-Z(\eta)\right] h(U),
$$
where $\eta_i \in \mathbb{R}$ ($i=1,\ldots,s$) are the parameters and $Z(\eta) \in \mathbb{R}$ is a function that maps the parameters to a real number, and the $T_i(x)$'s are known as the sufficient statistics. The term $Z(\eta)$ is known as the log-normalization constant or the log-partition function, which is meant to make the distribution's integral equal to 1, i.e.
$$
Z(\eta)=\log \left[\int \exp \left[\sum_{i=1}^s \eta_i T_i(U)\right] h(U) d U\right].
$$
Taking the derivatives of $Z$ with respect to $\eta$ we obtain that,
$$
\begin{aligned}
\frac{\partial Z(\eta)}{\partial \eta_i} & =\frac{\int_{\mathcal{X}} T_i(U) \exp \left[\sum_{i=1}^s \eta_i T_i(U)\right] h(U) d U}{\left[\int_{\mathcal{X}} \exp \left[\sum_{i=1}^s \eta_i T_i(U)\right] h(U) d U\right]} \\
& =\mathbb{E}\left[T_i(U)\right].
\end{aligned}
$$
Exponential family is a big class of distributions which has great expressive ability.
Next, we first give a demonstration of exponential family by showing that  quadratic total cost leads to a distribution in the exponential family. Then, we use exponential family to generalize beyond the quadratic case in \cref{inf:lip}, which is the main result of this section. 

\begin{example}[Exponential Family for Quadratic Case] Consider the quadratic case where $J(U) =U^\top D U + d^\top U = (U-U^*)^\top D (U-U^*) + J^*$ . The normal distribution and the exponential inverse distribution of the total cost $P(U)\propto \mathcal{N}(U\mid U_{\mathrm{in}},\Sigma)$, $P(U) \propto \exp(-\frac{J(U)}{\lambda})$ can both be expressed in the form of an exponential family distribution. To see this, consider an exponential family: 
$$
P(U;\eta_1,\eta_2)=h(U) \exp(\eta_1^\top U + U^\top \eta_2 U -Z(\eta_1,\eta_2)),
$$
where \(\eta_1 \in \mathbb{R}^{mH},\eta_2 \in \mathbb{R}^{mH\times mH}\) are the natural parameters of the family. \(Z(\eta_1,\eta_2)\) is the log-partition of the exponential family, ensuring the integral of the distribution equals $1$. The sufficient statistics are set as $T_1(U) = U$ and $T_2(U) = U U^\top$. Here $T_2(U)\in \mathbb{R}^{mH\times mH}$ and it's straight forward that $U^\top \eta_2 U = \sum_{i=1}^{mH}\sum_{j=1}^{mH} [\eta_2]_{ij} [U U^\top]_{ij}$, where $[\quad]_{ij}$ denote the $i$'th row and $j$'th column's element of the matrix. Note that 
\begin{align*}
    P(U)&\propto \mathcal{N}(U\mid U_{\mathrm{in}},\Sigma) \propto \exp(-\frac{1}{2}(U-U_\mathrm{in})^\top \Sigma^{-1}(U-U_\mathrm{in})) \propto \exp(U_{\mathrm{in}}^\top \Sigma^{-1} U + U^\top(- \frac{1}{2}\Sigma^{-1})U).
\end{align*}
We can then tell that \(\mathcal{N}(U_{\mathrm{in}},\Sigma)\) follows a distribution from this family with \(\eta_1 = \Sigma^{-1}U_{\mathrm{in}},\eta_2 = -\frac{1}{2}\Sigma^{-1}\).

Likely, the distribution \(P(U)\propto \mathcal{N}(U\mid U_{\mathrm{in}},\Sigma)\exp(-\frac{J(U)}{\lambda})\) is also part of this exponential family, because:
\begin{align*}
P(U)&\propto \mathcal{N}(U\mid U_{\mathrm{in}},\Sigma)\exp(-\frac{J(U)}{\lambda}) \propto \exp(U_{\mathrm{in}}^\top \Sigma^{-1} U + U^\top(- \frac{1}{2}\Sigma^{-1})U)\exp(-\frac{1}{\lambda}(U^\top D U+U^\top d))  \\
 &\propto  \exp((U_{\mathrm{in}}^\top \Sigma^{-1}-\frac{1}{\lambda} d^\top) U + U^\top (-\frac{1}{2}\Sigma^{-1}-\frac{1}{\lambda}D)U) \\
 & \propto \exp((U_{\mathrm{in}}^\top \Sigma^{-1}-\frac{2}{\lambda} U^{*\top}D) U + U^\top (-\frac{1}{2}\Sigma^{-1}-\frac{1}{\lambda}D)U)
\end{align*}
where the last $\propto$ comes from the variant of $J(U)$ that $J(U) = U^\top D U + d^\top U = (U-U^*)^\top D (U-U^*)$, and the second equivalent comes from a constant level shift of $J$ does not affect the result. Therefore, $P(U) \propto \mathcal{N}(U\mid U_{\mathrm{in}},\Sigma)\exp(-\frac{J(U)}{\lambda})$ also follows a distribution in the exponential family with parameters \(\eta_1 =  \Sigma^{-1} U_{\mathrm{in}}-\frac{1}{\lambda} d,\eta_2 = -\frac{1}{2}\Sigma^{-1}-\frac{1}{\lambda}D\) or $\eta_1 = \Sigma^{-1}U_{\mathrm{in}} -\frac{2}{\lambda} D U^{*}, \eta_2 = -\frac{1}{2}\Sigma^{-1}-\frac{1}{\lambda}D$.

Recall that, according to \eqref{eq:epu}, we can tell that $U_{\mathrm{out}} \xrightarrow[p]{} \mathbb{E}_{P}\mathcal{U}$, where $ P(U)\propto \mathcal{N}(U\mid U_{\mathrm{in}},\Sigma)\exp(-\frac{J(U)}{\lambda}) $. Leveraging the characteristics of the exponential family, we have $\mathbb{E}_{P}\mathcal{U} = \mathbb{E}_P(T_1(\mathcal{U})) = \frac{\partial Z(\eta)}{\partial \eta_1}$. 
\end{example}

With the help of the exponential family, we demonstrate that even with bounded residual nonlinear dynamics (in which case $J(U)$ is no longer quadratic), contraction is still guaranteed to some extent. Specifically, we consider the same cost function as in \cref{eq:cost_func} with the following dynamics: $x_{t+1} = A x_t+B [U]_t + g(x_t,[U]_t)$, where \(g\) represents a small residual nonlinear dynamics. 

\begin{statement}{Theorem 5}
Suppose the log-partition function's derivative $\frac{\partial Z(\boldsymbol{\eta})}{\partial \eta_1}$  is  $L_1$-lipschitz continuous, the optimal control under the residual dynamic $U^{*'}$ and under the original dynamic $U^*$  are both bounded by a constant that $\left\| U^{*}\right\| \leq c_{u*}$ and $\left\| U^{*'}\right\| \leq c_{u*}$, and the residual dynamics $g$ satisfies some mild assumption that $\left\|g(x,u)\right\|\leq c_0,\left\|\frac{\partial g}{\partial x}\right\|\leq c_{x1} \leq \min(\frac{A}{H},(\left\|A\right\| -1)\left\|A\right\|),\left\|\frac{\partial g}{\partial U}\right\|\leq c_{u1}, \left\|\frac{\partial^2 g}{\partial U \partial x}\right\|\leq c_{xu}, \left\|\frac{\partial^2 g}{\partial x^2 }\right\|\leq c_{x2}$. We further assume that the optimal trajectory ${x_i^*}$ is also bounded that $\left\|x_i^*\right\| \leq c_{x*}$. Then,
    \begin{equation}
        U_{\mathrm{out}} - U^* \xrightarrow[p]{} (I-\frac{2}{\lambda}(\Sigma^{-1}+\frac{2}{\lambda}D)^{-1}D^\top)(U_{\mathrm{in}} - U^*) + U_{\mathrm{error}},
    \end{equation}
    where $\left\|U_{\mathrm{error}}\right\|\sim \frac{L_1}{\lambda}\left\| Q \right\|(O(\left\| A^{3H}\right\|)C_3 + O(\left\|A^{2H}\right\|)C_2)$, with $C_2 = (1+2 c_{u*})4\left\|B\right\| \left\|A\right\| $ $(c_{u1} + H c_{x1} \frac{\left\|AB\right\|+c_{u1}}{\left\|A\right\|-c_{x1}H} ) +  2  (c_{u1}+H c_{x1}\frac{\left\|AB\right\|+c_{u1}}{\left\|A\right\|-c_{x1}H})^2 + 2 c_{x*}  \left\|A\right\| (\frac{\left\|AB\right\|+c_{u1}}{\left\|A\right\|-c_{x1}H})^2 (\frac{c_{x2}}{(\left\|A\right\| -1)\left\|A\right\|-c_{x1}}), $ 
    and
    $C_3 = (1+2 c_{u*})c_0  (\frac{\left\|AB\right\|+c_{u1}}{\left\|A\right\|-c_{x1}H})^2 (\frac{c_{x2}}{(\left\|A\right\| -1)\left\|A\right\|-c_{x1}})$
    \label{thm:lip_continuious}
\end{statement}
\begin{proof}
In the proof, we focus on 1-dimensional system that $A, B, Q\in \mathbb{R}$ are scalars where $x_t, [U]_t \in \mathbb{R}$ are also scalars. This is for the purpose of simplifying notation can can easily generalize to the multi-dimensional case.
Here we denote $J_{\mathrm{wr}}(U)$ as the total cost with the residual dynamics, and $x^{\mathrm{res}}$ is the state with the residual dynamics. In other words,

\begin{align*}
          J_{\mathrm{wr}}\left(U\right) &= \sum_{h=1}^H x_h^{\mathrm{res} \top} Q x_h^{\mathrm{res}} + [U]_h^\top R [U]_h \\ \text {with dynamics} \quad x_{h+1}^{\mathrm{res}} &= A x_h^{\mathrm{res}} + B [U]_h + g(x_h^{\mathrm{res}},[U]_h), \quad 1 \leq h \leq H. 
\end{align*}
The counterpart to this is the nominal trajectory total cost, which we now denote as $J_{\mathrm{n}}(U)$, and $x_{\mathrm{n}}$ respectively.
\begin{align*}
          J_{\mathrm{n}}\left(U\right) & = \sum_{h=1}^H x_h^{\mathrm{nom}\top} Q x_h^{\mathrm{nom}} + [U]_h^\top R [U]_h  \\  \text {with dynamics}\quad & x_{h+1}^{\mathrm{nom}}  = A x_h^{\mathrm{nom}} + B [U]_h, \quad 1 \leq h \leq H
\end{align*}
For the simplicity of further derivation, we define $\delta x_{h} = x_h^{\mathrm{res}} - x_h^{\mathrm{nom}}$. Therefore, 
$$
J_{\mathrm{wr}}\left(U\right) = \sum_{h=1}^H (x_h^{\mathrm{nom}} +\delta x_h)^\top Q (x_h^{\mathrm{nom}} +\delta x_h) + [U]_h^\top R [U]_h .
$$
We also have $J_{\mathrm{n}}(U) = (U-U^*)^\top D (U-U^*)$. Further, we denote the optimal point of $J_{\mathrm{wr}}$ as $U^{*'}$. Then, we can split $J_{\mathrm{wr}}$ into $J_{\mathrm{wr}}(U) = (U-U^{*'})^\top D' (U-U^{*'}) + J_{\mathrm{res}}(U)$, where $D'= \frac{\partial^2 J_{\mathrm{wr}}}{\partial U^2}\mid_{U=U^{*'}}$, $J_{\mathrm{res}}(U) = J_{\mathrm{wr}}(U)-(U-U^{*'})^\top D' (U-U^{*'})$ is a higher order residual cost with $\frac{\partial J_{\mathrm{res}}}{\partial U}\mid_{U=U^{*'}} = 0, \frac{\partial^2 J_{\mathrm{res}}}{\partial U^2}\mid_{U=U^{*'}}=0$.

The remaining of the proof is divided into two parts. In the first part, we show that $U_{\mathrm{out}}$ converges to the expectation of an exponential family distribution. We further show that by decomposing the total cost, the difference in the expectation can be bounded by the difference of Hessian brought by the residual dynamics. For the second part, we show that for small enough residuals with bounded derivatives, the difference of the Hessian matrix can be bounded. 

\noindent\textbf{Part 1: Bound the Expectation by Lipschitz of Log-Partition.}
In this proof, we first augment the exponential family with an additional sufficient statistics $T_3(x) = J_{\mathrm{res}}(x)$. The exponential family can 
now be denoted as:
$$
P(U;\eta_1,\eta_2,\eta_3)=h(x) \exp(\eta_1^\top U + U^\top \eta_2 U + \eta_3 J_{\mathrm{res}}(U) -Z(\eta_1,\eta_2,\eta_3)).
$$
We first consider a distribution $P$ from the exponential family that $\eta_1 =  \Sigma^{-1}U_{\mathrm{in}}-\frac{2}{\lambda} DU^{*'},\eta_2 = -\frac{1}{2}\Sigma^{-1}-\frac{1}{\lambda}D, \eta_3 =0 $. From the derivation in \cref{theo:1_contraction}, the expectation of $\mathbb{E}_P(\mathcal{U})$ on this distribution is 
$$\mathbb{E}_P(\mathcal{U}) = (I-\frac{2}{\lambda}(\Sigma^{-1}+\frac{2}{\lambda}D)^{-1}D^{\top})(U_{\mathrm{in}} - U^{*'}).$$
However, $U_{\mathrm{out}}$ under the residual dynamics converge to the expectation of the distribution 
$$
P_{\mathrm{wr}}(U) \propto \exp(-\frac{1}{\lambda} J_{\mathrm{wr}}(U))\mathcal{N}(U\mid U_{\mathrm{in}},\Sigma),
$$
which is a distribution in the exponential family with slightly different natural parameters $\eta_1' = \Sigma^{-1} U_{\mathrm{in}}-\frac{2}{\lambda} D' U^{*'},\eta_2' = -\frac{1}{2}\Sigma^{-1}-\frac{1}{\lambda}D', \eta_3' =-\frac{1}{\lambda}$.

With the Lipschitz constant of $\frac{\partial Z(\boldsymbol{\eta})}{\partial \eta_1}$, we can bound the difference of $\mathbb{E}_{P}(\mathcal{U}) $ and $\mathbb{E}_{P_{\mathrm{wr}}}(\mathcal{U})$ as follows  
\begin{align}
\left\|\mathbb{E}_{P}(\mathcal{U})-\mathbb{E}_{P_{\mathrm{wr}}}(\mathcal{U})\right\|& \leq L_1(\left\|\eta_1-\eta_1'\right\| +\left\|\eta_2-\eta_2'\right\|+\left\|\eta_3-\eta_3'\right\|) 
\nonumber
\\
&=L_1(\frac{2}{\lambda}\left\|U^{*'\top}(D-D') \right\|+\frac{1}{\lambda}\left\|D-D'\right\|) 
\nonumber
\\
& \leq L_1(\frac{1}{\lambda}(2\left\|U^{*'\top}\right\|+1)\left\|D-D'\right\|) \\
& \leq L_1(\frac{1}{\lambda}(2 c_{u*}+1)\left\|D-D'\right\|).
\label{eq:lip_hessian}
\end{align}

From above, one can see the remaining task is to bound $\Vert D-D'\Vert$, which we do now.

\bigskip
\noindent\textbf{Part 2: Bounding the difference of Hessian brought about by the residual dynamics.}
The trajectory after adding the residual can be calculated as  
$$
x_t^{\mathrm{res}} = A^t x_0^{\mathrm{res}} + \sum_{i=0}^{t-1} A^{t-i} B U_i + \sum_{i=0}^{t-1}  A^{t-i-1} g(x^{\mathrm{res}}_i,[U]_i).
$$
Meanwhile, the nominal trajectory can be calculated as 
$$
x_t^{\mathrm{nom}} = A^t x_0^{\mathrm{nom}} + \sum_{i=0}^t A^{t-i} B U_i.
$$
The first order derivative of the trajectory with respect to $U$ can be denoted as 
\begin{equation}
\frac{\partial x_t^{\mathrm{res}}}{\partial [U]_k} =  A^{t-k}B +  A^{t-k-1}\frac{\partial g}{\partial [U]_k} + \sum_{i=k+1}^{t-1} A^{t-i-1} (  \frac{\partial g}{\partial x_i^{\mathrm{res}}} \frac{\partial x_i^{\mathrm{res}}}{\partial [U]_k} ).
\label{eq:x_u1_bound}
\end{equation}
    We next prove by induction that $\frac{\partial x_i^\mathrm{res}}{\partial [U]_j} \leq  \alpha_1 \left\| A^{i-j}\right\| $ holds for some $\alpha_1$ to be determined later. Assuming this holds for a pair of $i,j$, and substituting it into \eqref{eq:x_u1_bound}, we can find that  
    \begin{align*}
        \left\|\frac{\partial x_t^{\mathrm{res}}}{\partial [U]_k}\right\| & = \left\| A^{t-k}B +  A^{t-k-1}\frac{\partial g}{\partial [U]_k} + \sum_{i=k+1}^{t-1} A^{t-i-1} (  \frac{\partial g}{\partial x_i^{\mathrm{res}}} \frac{\partial x_i^{\mathrm{res}}}{\partial [U]_k} )\right\| \\
        & \leq \left\| A^{t-k}B \right\|+  \left\|A^{t-k-1}\frac{\partial g}{\partial [U]_k}\right\| + \sum_{i=k+1}^{t-1}\left\|  A^{t-i-1}(\frac{\partial g}{\partial x_i^{\mathrm{res}}} \frac{\partial x_i^{\mathrm{res}}}{\partial [U]_k} )\right\| \\
        & \leq \left\|A^{t-k}B\right\| +  c_{u1}\left\|A^{t-k-1} \right\| + c_{x1}\alpha_1 H \left\|A^{t-k-1}\right\|.
    \end{align*}
    The Last inequality comes from the induction assumption and the truth that $t-k-1\leq H$.
    
For small residual dynamics $g$, as aforementioned, we assume $c_{x1}\leq \frac{1}{H}\left\|A\right\|$. Therefore, when $\alpha_1 \geq \frac{\left\|AB\right\|+c_{u1}}{\left\|A\right\|-c_{x1}H}$, we have $\frac{\partial x_i^\mathrm{res}}{\partial [U]_j} \leq  \alpha_1 \left\| A^{i-j}\right\| $ holds for every $0 \leq i,j \leq H$.
    Similarly 
    $$
    \left\|\frac{\partial \delta x_t}{\partial [U]_k}\right\| =   \left\| A^{t-k-1} c_{u1} + \sum_{i=k+1}^{t-1} A^{t-i-1} (  c_{x1}  \alpha_1  A^{i-k} )\right\| \leq (c_{u1}+(t-k-1)c_{x1}\alpha_1)\left\|A^{t-k-1}\right\|.
    $$
    Then $\frac{\partial \delta x_t}{\partial [U]_k}$ is bounded by $\left\|\frac{\partial \delta x_t}{\partial [U]_k}\right\| \leq \alpha_{\delta 1}\left\|A^{t-k-1}\right\|$, where $\alpha_{\delta 1} = c_{u1}+H c_{x1}\alpha_1$.
    Likely, the second order derivative can be calculated: $$
    \frac{\partial^2 x_t^\mathrm{res}}{\partial [U]_k [U]_l} =A^{t-k-1}\frac{\partial^2 g}{\partial [U]_k \partial x_k }\frac{\partial x_k^\mathrm{res}}{\partial [U]_l}  + \sum_{i=l+1}^{t-1}  A^{t-i-1} (\frac{\partial^2 g}{\partial x_i^2} \frac{\partial x_i^\mathrm{res}}{\partial [U]_k} \frac{\partial x_i^\mathrm{res}}{\partial [U]_l} +  \frac{\partial g}{\partial x_i} \frac{\partial^2 x_i^\mathrm{res}}{\partial [U]_k \partial [U]_l}).
    $$
    Without loss of generality, here we assume $l \leq k$. Because the residual dynamic is bounded $\left\|\frac{\partial^2 g}{\partial [U]_k \partial x_k } \right\|\leq c_{xu}, \left\|\frac{\partial^2 g}{\partial x_i^2} \right\|\leq c_{x2}$ and this leads to, 

    \begin{align}
        \left\|\frac{\partial^2 x_t^\mathrm{res}}{\partial [U]_k [U]_l} \right\| & \leq \alpha_1 c_{xu} \left\|A^{t-l-1}\right\|  + \left\|\sum_{i=l+1}^{t-1}c_{x2}  \alpha_1^2 A^{t+i-k-l-1}  + c_{x1 }\frac{\partial^2 x_i^\mathrm{res}}{\partial [U]_k \partial [U]_l} A^{t-i-1} \right\| \nonumber\\ 
        &= c_{xu}\alpha_1 \left\|A^{t-l-1}\right\|  + c_{x2}  \alpha_1^2 \left\|\frac{A^{2t-k-l-1}}{(\left\|A\right\| -1)} \right\| +\sum_{i=l+1}^{t-1}c_{x1 } \left\|A^{t-i-1}\right\|\left\|\frac{\partial^2 x_i^\mathrm{res}}{\partial [U]_k \partial [U]_l}\right\|
        \label{eq:x_u_2}
    \end{align}
    Again, we bound $\left\|\frac{\partial^2 x_t^\mathrm{res}}{\partial [U]_k [U]_l} \right\|$ by induction method. Assuming $\left\|\frac{\partial^2 x_t^\mathrm{res}}{\partial [U]_k [U]_l} \right\| \leq \alpha_2 \left\|A^{2t-k-l}\right\|$, by the assumption and \eqref{eq:x_u_2}, we show that 
    $$\left\|\frac{\partial^2 x_t^\mathrm{res}}{\partial [U]_k [U]_l} \right\| \leq \alpha_1 c_{xu} \left\|A^{t-l-1}\right\|  + c_{x2}  \alpha_1^2 \left\|\frac{A^{2t-k-l-1}}{(\left\|A\right\| -1)}\right\| + c_{x1 }\alpha_2 \left\|\frac{A^{2t-k-l-1}} {(\left\|A\right\| -1)}\right\|.$$  With the assumption on $c_{x1} \leq (\left\|A\right\| -1)\left\|A\right\|$ from property of residual dynamic, we further get $\left\|\frac{\partial^2 x_t^\mathrm{res}}{\partial [U]_k [U]_l}\right\| \leq \alpha_2 \left\|A^{2t-k-l-1}\right\|$ holds by induction when $$\alpha_2 \geq     \frac{c_{x2} \alpha_1^2}{(\left\|A\right\| -1)\left\|A\right\|-c_{x1}} =     (\frac{\left\|AB\right\|+c_{u1}}{\left\|A\right\|-c_{x1}H})^2  \frac{c_{x2}}{(\left\|A\right\| -1)\left\|A\right\|-c_{x1}}.$$ Notice that $\frac{\partial^2 x_t^\mathrm{nom}}{\partial [U]_k [U]_l} = 0$, we can conclude that  $\left\|\frac{\partial^2 \delta x_t}{\partial [U]_k [U]_l}\right\| =  \left\|\frac{\partial^2 x_t^\mathrm{res}}{\partial [U]_k [U]_l}\right\|\leq \alpha_2 \left\|A^{2t-k-l-1}\right\|$.
    With the above bound on $\frac{\partial^2 x_t}{\partial [U]_k [U]_l}$, and $\frac{\partial x_t}{\partial [U]_k}$, we now formulate the total cost $J$'s derivative $\frac{\partial^2 J}{\partial [U]_k [U]_l}$. The first-order derivative can be calculated as 
    
    $$
    \frac{\partial J}{\partial [U]_k} = 2R[U]_k  + \sum_{i=k+1}^{H} 2 x_i^\top Q \frac{\partial x_i}{\partial [U]_k}.
    $$
    And Assuming $l \geq k$, the second order derivative of the total cost $J$ can be formulated as:
    $$
    \frac{\partial^2 J}{\partial [U]_k [U]_l} =\sum_{i=l+1}^{H} 2 \frac{\partial x_i}{\partial [U]_k}^\top Q \frac{\partial x_i}{\partial [U]_l} + 2 x_i^\top Q \frac{\partial^2 x_i}{\partial [U]_k \partial [U]_l}.
    $$
    Here we define the cost under the residual dynamics as $J_{\mathrm{ws}}$ and the cost under the nominal trajectory as $J_{\mathrm{n}}$
    And the difference between $\frac{\partial^2 J_{\mathrm{ws}}}{\partial [U]_k^2}$ and $\frac{\partial^2 J_{\mathrm{n}}}{\partial [U]_k^2}$ can be calculated:

    \begin{align*}
        \frac{\partial^2 J_{\mathrm{ws}}}{\partial [U]_k \partial [U]_l}-\frac{\partial^2 J_{\mathrm{n}}}{\partial [U]_k\partial [U]_l}  &= \sum_{i=k+1}^{H} 2\frac{\partial x_i}{\partial [U]_k}^\top Q \frac{\partial \delta x_i}{\partial [U]_l} + 2\frac{\partial x_i}{\partial [U]_l}^\top Q \frac{\partial \delta x_i}{\partial [U]_k}+2\frac{\partial \delta x_i}{\partial [U]_k}^\top Q \frac{\partial \delta x_i}{\partial [U]_l} \\
        & + 2 x_i^\top Q \frac{\partial^2 (\delta x_i)}{\partial [U]_k\partial [U]_l} + 2 \delta x_i^\top Q(\frac{\partial^2 (\delta x_i)}{\partial [U]_k\partial [U]_l}).
    \end{align*}
    With a slight abuse of notation, we want to stress that $x$ denote the trajectory under the residual dynamic $g$, $\delta x = x - x^{\mathrm{nom}}$ denote its difference to the nominal trajectory $x^{\mathrm{nom}}$, 
    Substituting $\frac{\partial x}{\partial [U]}$, $\frac{\partial \delta x}{\partial [U]}$,$\frac{\partial^2 \delta x}{\partial [U]^2}$ with their bound, we show that
    \begin{align*}
        &\left\|\frac{\partial^2 J_{\mathrm{ws}}}{\partial [U]_k \partial [U]_l}-\frac{\partial^2 J_{\mathrm{n}}}{\partial [U]_k\partial [U]_l}\right\|  & \\
        & \leq \sum_{i=k+1}^{H} \left\|2\frac{\partial x_i}{\partial [U]_k}^\top Q \frac{\partial \delta x_i}{\partial [U]_l}\right\| + \left\|2\frac{\partial x_i}{\partial [U]_l}^\top Q \frac{\partial \delta x_i}{\partial [U]_k}\right\| +\left\| 2\frac{\partial \delta x_i}{\partial [U]_k}^\top Q \frac{\partial \delta x_i}{\partial [U]_l}\right\| \\ 
        & + \left\|2 x_i^\top Q \frac{\partial^2 (\delta x_i)}{\partial [U]_k\partial [U]_l}\right\| + \left\| 2 \delta x_i^\top Q \frac{\partial^2 (\delta x_i)}{\partial [U]_k\partial [U]_l}\right\|& \\
        & \leq \sum_{i=k+1}^{H}( \alpha_{\delta 1}\left\| A^{i-k}B\right\| \left\|Q\right\|  \left\|A^{i-l-1}\right\| + 2\alpha_{\delta 1}\left\|  A^{i-l}B\right\|\left\|Q\right\| \left\| A^{i-k-1} \right\| + 2 \alpha_{\delta 1}^2 \left\|A^{i-k-1}\right\| \left\|A^{k-l-1}\right\| \left\| Q \right\| \\
        & + 2c_{x*} \alpha_2 \left\|  Q \right\| \left\| A^{2i-k-l-1}\right\| +  \left\| \sum_{j=0}^{i-1} A^{i-j-1} g(x^{\mathrm{res}}_i,[U]_i)\right\| \left\| Q\right\| \left\| \alpha_2 A^{2i-k-l-1}\right\| \\
        & \leq \sum_{i=k+1}^{H}(\left\|A^{2i-k-l-2}\right\|(4 \alpha_{\delta 1}\left\|Q\right\| \left\|B\right\| \left\|A\right\| + 2 \left\|Q\right\| \alpha_{\delta_1}^2 + 2 c_{x*} \left\|Q\right\| \alpha_2 \left\|A\right\|) + c_0 \left\|Q\right\| \alpha_2 \sum_{j=0}^{i-1}  \left\|A^{i-j}\right\|) \\
        & \leq \left\|\frac{A^{2H-k-l}-A^{k-l}}{A^2-1}\right\| (4 \left\|B\right\| \left\|Q\right\| \left\|A\right\| \alpha_{\delta_1}+2 \left\|Q\right\| \alpha_{\delta_1}^2 + 2  \alpha_2 c_{x*} \left\| Q \right\|  \left\|A\right\|)\\ 
        &  + c_0 \alpha_2  \left\|Q\right\| \left\|\frac{A^{3H-k-l+2}-A^{2k-l+2}}{A^3-1}\right\|.
    \end{align*}
    Where the trajectory is bounded by $\left\| x_t \right\| \leq c_{x*}$ and residual is bounded by $\left\|g(x,u)\right\|\leq c_0$.
    Therefore, with the beginning assumption that the system is 1-d, we can get that
    \begin{align*}
         \sum_{k=1}^H \sum_{l=1}^H \left\|\frac{\partial^2 J_{\mathrm{ws}}}{\partial [U]_k \partial [U]_l}-\frac{\partial^2 J_{\mathrm{n}}}{\partial [U]_k\partial [U]_l}\right\|  &\leq O(\left\| A^{2H}\right\|)(4\left\|B\right\|\left\|Q\right\|\left\|A\right\| \alpha_{\delta 1}+ 2 \left\|Q\right\| \alpha_{\delta 1}^2 + 2 \left\|Q\right\| \left\|A\right\| c_{x*}  \alpha_2 ) \\ 
         &  + O(\left\| A^{3H}\right\|)c_0 \left\|Q\right\| \alpha_2.
    \end{align*}
    Substituting $\alpha_{\delta 1}, \alpha_2$ into the right hand side. We can get :$ \mathrm{RHS} =  O(\left\| A^{2H}\right\|)(4\left\|B\right\|\left\|Q\right\|\left\|A\right\|    (c_{u1}+H c_{x1}\alpha_1) + 2 \left\|Q\right\| (c_{u1}+H c_{x1}\alpha_1)^2 + 2 c_{x*} \left\|Q\right\| \left\|A \right\|(\frac{\left\|AB\right\|+c_{u1}}{\left\|A\right\|-c_{x1}H})^2 \frac{c_{x2}}{(\left\|A\right\| -1)\left\|A\right\|-c_{x1}} + O(\left\| A^{3H}\right\|)c_0 \left\|Q\right\| $ $(\frac{\left\|AB\right\|+c_{u1}}{\left\|A\right\|-c_{x1}H})^2 \frac{c_{x2}}{(\left\|A\right\| -1)\left\|A\right\|-c_{x1}}$.
    
\end{proof}

\begin{remark}
   Consider the scenario of a constant residual. In this case, the Lipschitz constants for the derivatives of the residual dynamics \( g \) are all equal to zero. This implies that \( U_{\mathrm{error}} \) is also zero, further indicating that a constant residual does not introduce any error in convergence.
\end{remark}

\subsubsection{The General Case}
\label{apd:proof_general}
A simple but straightforward following corollary would be when we have direct access to the cost function's structure, i.e. knowing both of $D$ and $D'$ or the difference $J_{\mathrm{wr}}-J_{\mathrm{n}}$. Also, Notice that the Lipschitz constant $L_1$ is task-specific because different tasks will lead to sufficient statistics $J_{\mathrm{wr}}-J_{\mathrm{n}}$ different in the exponential family.

\begin{statement}{Corrollary 6} If the log-partition function's derivative $\frac{\partial Z(\boldsymbol{\eta})}{\partial \eta_1}$  is  $L_1$-lipschitz continuous. We denote the cost function under the residual dynamics as $J_{\mathrm{wr}}(U)$, the cost function of the nominal dynamics as  $J_{\mathrm{n}} = U^\top D U + d^\top U$, and the optimal control under the residual as $U^{*'}$. Then
    \begin{equation}
        U_{\mathrm{out}} - U^* \xrightarrow[p]{} (I-\frac{2}{\lambda}(\Sigma^{-1}+\frac{2}{\lambda}D)^{-1}D)(U_{\mathrm{in}} - U^{*'}) + U_{\mathrm{error}}.
    \end{equation}
    with $\left\|U_{\mathrm{error}}\right\|\sim O(\frac{L_1}{\lambda}\left\|D - \frac{\partial^2 J_{\mathrm{wr}}}{\partial U^2}\mid_{U=U^{*'}} \right\|)$.
    \label{cor:general_non}
\end{statement}
\begin{proof}
    From \cref{eq:lip_hessian} in proof of Theorem \ref{inf:lip}, obviouslly, $U_{\mathrm{error}}$ can be bounded by the difference of Hessian, i.e. $\left\|D - \frac{\partial^2 J_{\mathrm{wr}}}{\partial U^2}\mid_{U=U^{*'} } \right\|$.
\end{proof}

\section{Implementation Details}

\subsection{Algorithm Implementation}
\label{sec:algo_impl}

The annotated pseudocode for the \covo algorithm is shown in \cref{alg:covo-appendix}. The full version of offline approximation \covo is shown in \cref{alg:covo-offline-appendix}. 
The major difference between those two algorithms is that the offline approximation uses the covariance matrix from the buffer instead of calculating it online. Here we describe the PID controller used in our offline approximation:

\label{sec:offline}
For the \texttt{Quadrotor} environment, we generate the linearization point using a differential-flatness-based nonlinear controller:

\begin{align*}
    \boldsymbol{a}_{\mathrm{fb}}       & =-K_{P}\left(\boldsymbol{p}-\boldsymbol{p}^{d}\right)-K_{D}\left(\boldsymbol{v}-\boldsymbol{v}^{d}\right)-K_{I} \int\left(\boldsymbol{p}-\boldsymbol{p}^{d}\right)+\boldsymbol{a}^{d}-\boldsymbol{g} \\
    \boldsymbol{z}_{\mathrm{fb}}       & =\frac{\boldsymbol{a}_{\mathrm{fb}}}{\left\|\boldsymbol{a}_{\mathrm{fb}}\right\|}, \quad \boldsymbol{z}=\boldsymbol{R e}_{3}, \quad f_{\Sigma}=\boldsymbol{a}_{\mathrm{fb}}^{\top} \boldsymbol{z}                             \\
    \boldsymbol{\omega}_{\mathrm{des}} & =-K_{R} \boldsymbol{z}_{\mathrm{fb}} \times \boldsymbol{z}+\psi_{\mathrm{fb}} \boldsymbol{z}, \quad \psi_{\mathrm{fb}}=-K_{\mathrm{yaw}}\left(\psi \ominus \psi_{\mathrm{ref}}\right)
\end{align*}

For the \texttt{Cartpole} environment, we use the linearization point from the simple feedback controller with gain \(K = \begin{bmatrix} 0.5 & 0.5 & 5.0 & 5.0 \end{bmatrix}\).

\begin{algorithm2e}[H]
    \label[algorithm]{alg:covo-appendix}
   
    \DontPrintSemicolon
    \LinesNumbered
    \SetAlgoLined
    \caption{CoVO-MPC: CoVariance-Optimal MPC}
    \KwIn{$H \gets \text{Horizon}$;
        $N \gets \text{Sample number};$
        $T \gets \text{Dynamic system time limit};$
        $x_0 \gets \text{Initial state};$
        $c_{1:T} \gets \text{Cost function};$
        $\mathbf{C}(\cdot) \gets \text{Optimal covariance function};$
        $\texttt{shift}(\cdot) \gets \text{Shift operator to 1 step forward};$}
    \BlankLine

    \For{$t=1:T$}{
    $\Sigma_{t} \gets \mathbf{C}(D_{t} = \nabla^2 J_t(U_{\mathrm{in}|t}))$ \tcp*{Calculate the sampling covariance matrix (our contribution here.)}
    $U_{i|t} \sim \mathcal{N}(U_{\mathrm{in}|t}, \Sigma_{t})$  \tcp*{Sample $N$ controls}
    $\kappa_{i|t} \gets \frac{\exp(-\beta J_t(U_{i|t}))}{\sum_{i=1}^N \exp(-\beta J_t(U_{i|t}))}$  \tcp*{Calculate the accumulated cost and Get the weight of each control}
    $U_{\mathrm{out}|t} \gets \sum_{i=1}^N \kappa_{i|t} U_{i|t}$  \tcp*{Calculate the future control sequence}
   
    Executing $[U_{\mathrm{out}|t}]_1$ to get $x_{t+1}$ \tcp*{Get the next control}
    
    $U_{\mathrm{in}|t+1} \gets \texttt{shift}(U_{\mathrm{out}|t})$ \tcp*{Update the mean with shift operator}

    }
\end{algorithm2e}

\begin{algorithm2e}[H]
    \label[algorithm]{alg:covo-offline-appendix}
    \DontPrintSemicolon
    \LinesNumbered
    \SetAlgoLined
    \caption{CoVO-MPC (offline approximation)}
    \KwIn{$H \gets \text{Horizon}$;
        $N \gets \text{Sample number};$
        $T \gets \text{Dynamic system time limit};$
        $x_0 \gets \text{Initial state};$
        $c_{1:T} \gets \text{Cost function};$
        $\mathbf{C}(\cdot) \gets \text{Optimal covariance function};$
        $\texttt{shift}(\cdot) \gets \text{Shift operator to 1 step forward};$}
    \BlankLine

    \tcp{Cache the covariance matrix for all future time step \(t=1:T\) with suboptimal controller $\pi_K$}
    \For{$t=1:T$}{
    \tcp{Rollout future step with nominal controller $\pi_K$}
    \For{$t'=1:H$}{
        $U_{\mathrm{in}}' \gets \pi_K(x_t')$ \tcp*{Get the next control}
        Executing $U_{\mathrm{in}}'$ to get $x_{t+1}'$

        Record control $[U_{\mathrm{off}|t}]_{t'} \gets U_{\mathrm{in}}$
    }
    $\Sigma_{\text{off}|t} \gets \mathbf{C}(D_{t} = \nabla^2 J_t(U_{\mathrm{off}|t}))$ \tcp*{Cache the covariance matrix}
        }

        \For{$t=1:T$}{
    $\Sigma_{t} \gets \Sigma_{\text{off}|t}$ \tcp*{Get covarance matrix from buffer}
    $U_{i|t} \sim \mathcal{N}(U_{\mathrm{in}|t}, \Sigma_{t})$  \tcp*{Sample $N$ controls}
    $\kappa_{i|t} \gets \frac{\exp(-\beta J_t(U_{i|t}))}{\sum_{i=1}^N \exp(-\beta J_t(U_{i|t}))}$  \tcp*{Calculate the accumulated cost and Get the weight of each control}
    $U_{\mathrm{out}|t} \gets \sum_{i=1}^N \kappa_{i|t} U_{i|t}$  \tcp*{Calculate the future control sequence}
    Executing $[U_{\mathrm{out}|t}]_1$ to get $x_{t+1}$ \tcp*{Get the next control}
    $U_{\mathrm{in}|t+1} \gets \texttt{shift}(U_{\mathrm{out}|t})$ \tcp*{Update the mean with shift operator}

    }
\end{algorithm2e}

For all the experiments, we keep the hyperparameter the same across all the experiments. 
We also keep the determinant of the covariance matrix the same between \covo and \mppi to make sure the sampling volume is the same.
The related hyperparameters are listed in \cref{tab:parameters}. 

\begin{table}[h]
    \centering
    \begin{tabular}{cc}
        \textbf{Parameter}            & \textbf{Value} \\
        \hline \hline
        Horizon \(H\)                 & 32             \\
        Sampling Number \(N\)         & 8192           \\
        Temperature \(\lambda\)       & 0.01           \\
        Sampling Covariance Determinant \(\alpha\) & \(0.5^{32}\)         \\
    \end{tabular}
    \caption{MPC hyperparameters}
    \label{tab:parameters}
\end{table}

\subsection{Environment Details}
\label{sec:env_detail}

We use the following dynamic model for the \texttt{Quadrotor} environment:

\begin{align}
    \boldsymbol{x}                         & =
    \begin{bmatrix}
        \boldsymbol{p} \\
        \boldsymbol{v} \\
        \boldsymbol{R} \\
        \boldsymbol{\omega}
    \end{bmatrix},
    \boldsymbol{u}                                    =
    \begin{bmatrix}
        \frac{T}{m} \\
        \boldsymbol{\tau}
    \end{bmatrix},
    \dot{\boldsymbol{x}}                              = \boldsymbol{f}(\boldsymbol{x}, \boldsymbol{u})                                                                                        \\
    \boldsymbol{f}(\boldsymbol{x}, \boldsymbol{u}) & = \begin{bmatrix}
                                               \boldsymbol{v}                                                                \\
                                               \boldsymbol{e}_3 g + \boldsymbol{R} \boldsymbol{e}_3  \frac{T}{m}  + \boldsymbol{d}_p \\
                                               \boldsymbol{R} \boldsymbol{\omega}                                            \\
                                               \boldsymbol{J}^{-1} \left( \boldsymbol{\tau} + \boldsymbol{d}_\tau  - \boldsymbol{\omega} \times \boldsymbol{J} \boldsymbol{\omega} \right)
                                           \end{bmatrix}
    \label{eq:dynamics}
\end{align}
where \(\boldsymbol{p}, \boldsymbol{v} \in \mathbb{R}^3\) are the position and velocity, \(\boldsymbol{R} \in \text{SO}(3)\) is the rotation matrix, \(\boldsymbol{\omega} \in \mathbb{R}^3\) is the angular velocity, \(\boldsymbol{u} \in \mathbb{R}^4\) is the control input, \(\boldsymbol{d}_p \in \mathbb{R}^3\) is the disturbance on the position~\citep{shi2019neural,o2022neural}, \(\boldsymbol{d}_\tau \in \mathbb{R}^3\) is the disturbance on the torque, \(\boldsymbol{e}_3\) is the unit vector in the \(z\) direction, \(g\) is the gravitational acceleration, \(m\) is the mass, and \(\boldsymbol{J} \in \mathbb{R}^{3 \times 3}\) is the inertia matrix. The disturbance \(\boldsymbol{d}_p\) is sampled from a zero-mean Gaussian distribution with a standard deviation denoted by \(\sigma_d\).

For hardware validation, we implement our algorithm on the Bitcraze Crazyflie 2.1 platform~\citep{giernackiCrazyflieQuadrotorPlatform2017}. Concurrently, we leverage Crazyswarm aiding communication~\citep{preissCrazyswarmLargeNanoquadcopter2017}. The state estimator acquires position data from an external OptiTrack motion capture system, while orientation data is relayed back from the drone via radio. Regarding control mechanisms, an on-board, lower-level PI body-rate controller $\boldsymbol{\tau}=-K_{P}^{\omega}\left(\boldsymbol{\omega}-\boldsymbol{\omega}_{\mathrm{des}}\right)-K_{I}^{\omega} \int\left(\boldsymbol{\omega}-\boldsymbol{\omega}_{\mathrm{des}}\right)$ operates at a frequency of 500Hz. This works together with an off-board higher-level controller, sending out desired thrust $f_d$ and body rate $\omega_d$ at a frequency of 50Hz. All communications are established using a 2.4GHz Crazyradio 2.0.

Here is an example of real-world experiment setup, where the Crazyflie is trying to tracking a triangular trajectory.

\begin{figure}[h]
    \centering
    \includegraphics[width=0.5\linewidth]{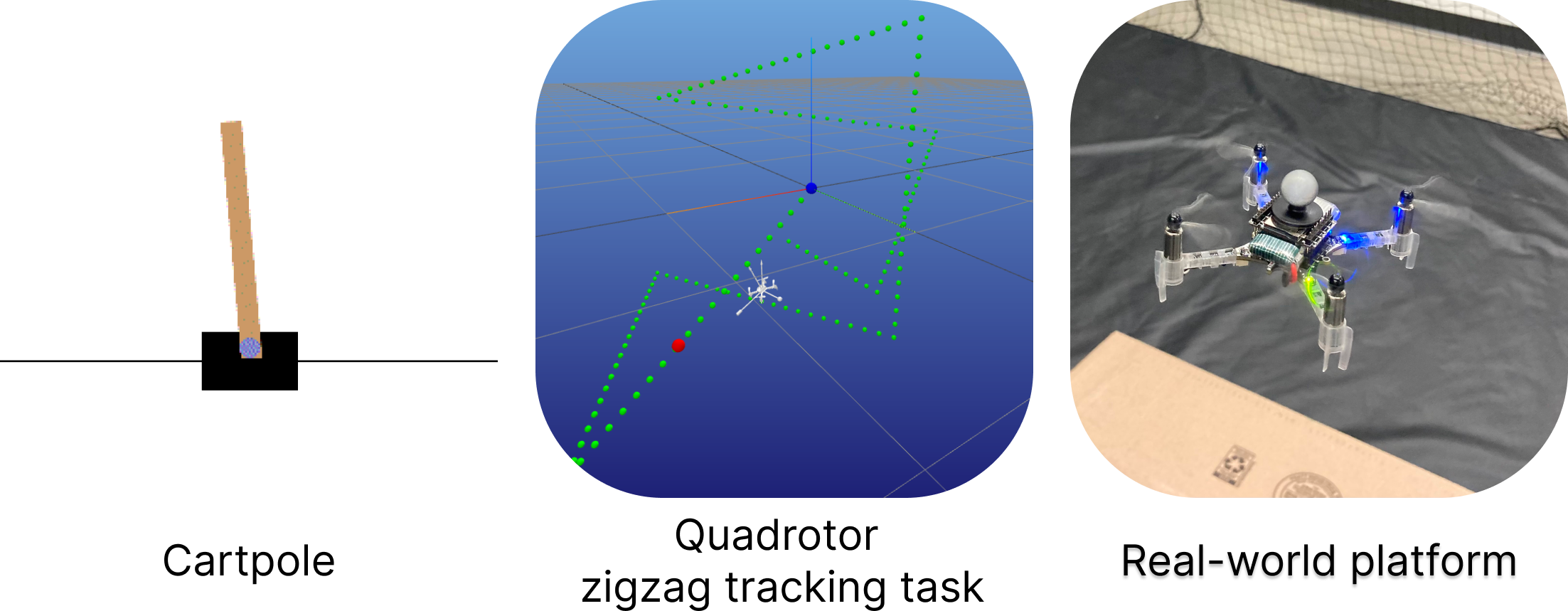}
    \caption{The experiment setup.}
    \label{fig:env}
\end{figure}

\section{Sampling distribution visualization}
\label{sec:dist_vis}

\cref{fig:cov_heatmap} shows the sampling covariance difference between two algorithms.
The right most one is the element-wise difference between \covo and \mppi. We can see that \covo generated a covariance matrix with richer patterns inlcuding:
\begin{enumerate}
    \item \textbf{Patterns inside a sigle control input}: \covo has a different scale for each control input, while \mppi has the same scale for all control inputs.
    \item \textbf{Patterns between control inputs}: \covo has a strong correlation between control inputs at different time steps, which also enables more effective sampling leveraging the dynamic system property.
    \item \textbf{Time-varying patterns}: \covo has a time-varying diagonal terms, which enable richer sampling schedule alongside the time-varying system dynamics.
\end{enumerate}

\begin{figure}
    \centering
    \includegraphics[width=\linewidth]{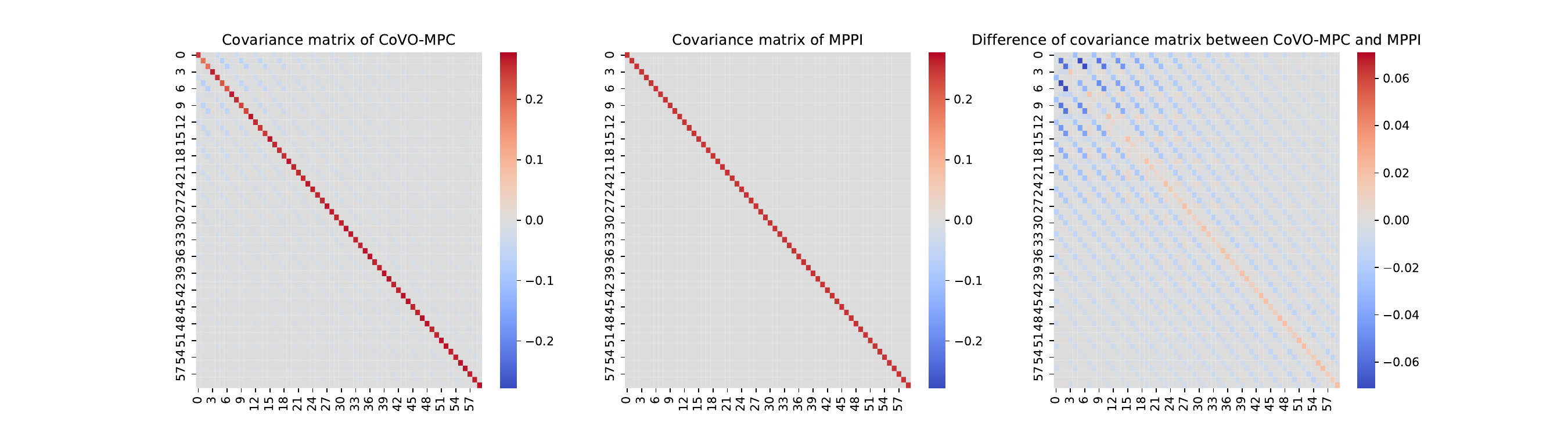}
    \caption{Covariance matrix visualization at certain timestep. The most right one is the element-wise difference between \covo and \mppi.}
    \label{fig:cov_heatmap}
\end{figure}
}{}

\end{document}